\documentclass[twocolumn,natbib]{svjour3}

\RequirePackage{fix-cm}
\usepackage{mathptmx}
\smartqed
\usepackage{microtype}

\usepackage{color}
\usepackage{xcolor}
\usepackage[normalem]{ulem}

\usepackage{graphicx}
\usepackage{commath}
\usepackage{multirow}
\usepackage{enumitem}

\usepackage{amsmath}
\usepackage{amsfonts}
\usepackage{amssymb}
\usepackage{mathtools}

\usepackage{hyperref}
\usepackage{bbm}
\usepackage{tabularx}

\newcommand{\R}{\mathbb R}
\newcommand{\Rdd}{\R^2}
\newcommand{\bp}{\bar\phi}
\newcommand{\bP}{\bar\Phi}
\newcommand{\M}{\mathbb M}
\newcommand{\Mi}{\mathbb M}
\newcommand{\W}{ W_s}
\newcommand{\om}{\Omega_\epsilon}
\newcommand{\omf}{\omega}
\newcommand{\E}{\mathbb E}
\newcommand{\Z}{\mathbb Z}
\newcommand{\C}{\mathbb C}
\newcommand{\Cov}{\mbox{Cov}}

\newcommand{\what}{\widehat}
\newcommand{\la}{\lambda}
\newcommand{\La}{\Lambda}
\newcommand{\Gaw}{\Gamma_H}
\newcommand{\jmax}{J}

\newcommand{\wxi} {{\xi_0}}
\newcommand{\rb}{\rangle}
\newcommand{\lb}{\langle}
\newcommand{\conv}{\star} 

\newcommand{\stkout}[1]{\ifmmode\text{\sout{\ensuremath{#1}}}\else\sout{#1}\fi}

\title{Particle gradient descent model for point process generation}

\author{Antoine Brochard \and Bart\l omiej B\l aszczyszyn \and Sixin Zhang \and St\'ephane Mallat}

\institute{A. Brochard \at
              Inria/ENS/PSL University, Paris, France \\
              \email{antoine.brochard@inria.fr}\\
              Huawei Technologies France, Boulogne-Billancourt, France
           \and
           B. B\l aszczyszyn \at
              Inria/ENS/PSL University, Paris, France
           \and
           S. Zhang \at
              IRIT, Universit\'e de Toulouse, CNRS, Toulouse, France
            \and
           S. Mallat \at
              ENS/PSL University, Paris, France\\
              Coll\`ege de France, Paris, France\\
              Flatiron Institute, New York, USA
}

\begin{document}

\maketitle

\begin{abstract}

This paper presents a statistical model for 
stationary ergodic point processes, estimated from a single realization observed in a square window.
With existing approaches in stochastic geometry, it is very difficult
to model processes with complex geometries formed by a large number of particles.
Inspired by recent works on gradient descent algorithms 
for sampling maximum-entropy models, 
we describe a model that allows 
for fast sampling of new configurations 
reproducing the statistics of the given observation. 
Starting from an initial random configuration, 
its particles are moved according to the gradient of an energy, in order to match a set of prescribed moments (functionals). 
Our moments are defined via a phase harmonic operator on the wavelet transform of point patterns. 
They allow one to
capture multi-scale interactions between the particles,
while controlling explicitly the number of moments 
by the scales of the structures to model. 
We present numerical experiments
on point processes with various geometric structures, and assess the quality of the model by spectral and topological data analysis.

\end{abstract}

\section*{Declarations}
\paragraph{Fundings}: This work was partly supported by the PRAIRIE 3IA Institute of the French ANR-19-P3IA-0001 program. Sixin Zhang was supported by the European Research Council (ERC FACTORY-CoG-6681839). Part of this work was done when Sixin Zhang was a postdoctoral researcher at ENS Paris, France.
\paragraph{Conflict of interest}The authors declare that they have no conflict of interest.
\paragraph{Availability of data and material} For the sake of transparency, we are ready to make available the data used to produce the results in our paper.
\paragraph{Code availability} For the sake of transparency, we are ready to make available the code used to produce the results in our paper.

\section{Introduction}

In order to generate new realizations of a stochastic process of which we have only one realization, 
we have to build a probabilistic model which approximates the distribution of this process, and from which we can sample.
In this article, 
we are interested in generative models for stationary, ergodic point processes. 
Such models are of interest in a wide range of applications \citep[Chapter 6]{illian}, for instance biology \citep{diggle2006modelling,baddeley2014multitype}, ecology \citep{ecology}, turbulent flows in atmosphere science \citep{PhysRevE77,PhysRevE.80.066312,MaOn19,oujia_matsuda_schneider_2020}, or cosmology \citep{cosmo2,tempel2016bisous}.
In some of these domains, the observed patterns exhibit complex structures, with a large number of particles (such as filaments in cosmology, or vortexes in turbulent flows). 
Our work is motivated by the simulation of such processes.

\begin{figure*}[!t]
    \centering
    \includegraphics[width=0.23\linewidth]{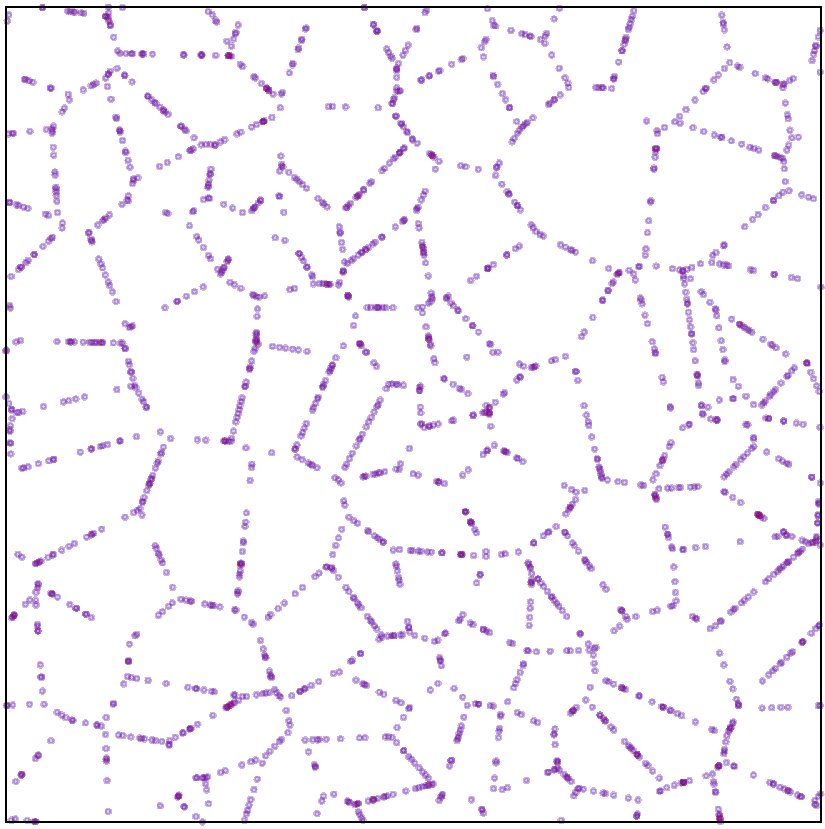}
    \includegraphics[width=0.23\linewidth]{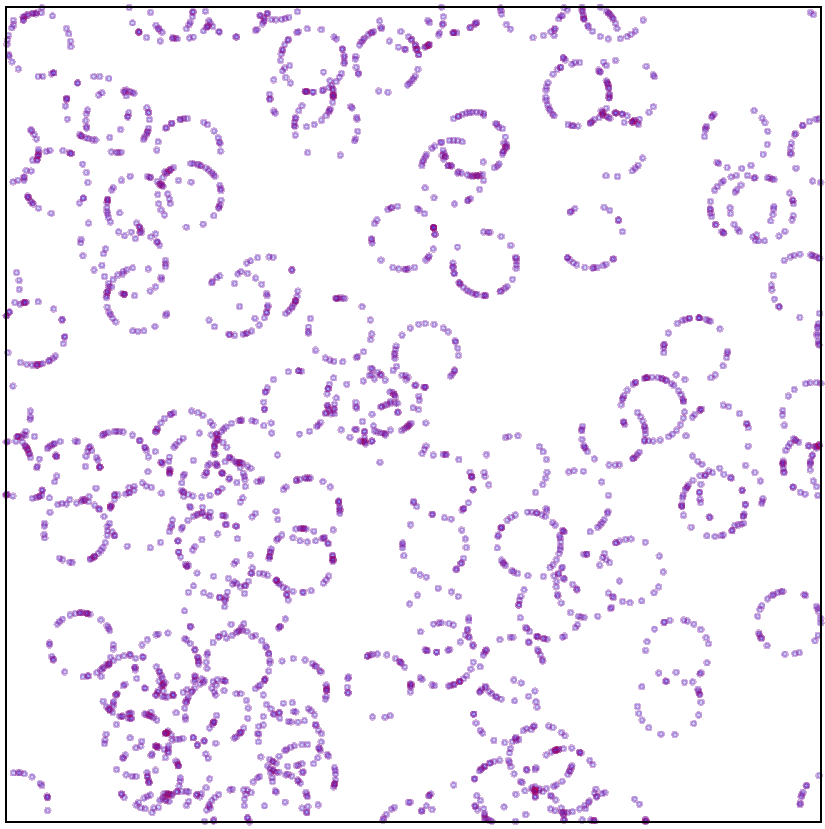}
    \includegraphics[width=0.23\linewidth]{Figures/Original_data/turb_zoom_sparse}
    \includegraphics[width=0.23\linewidth]{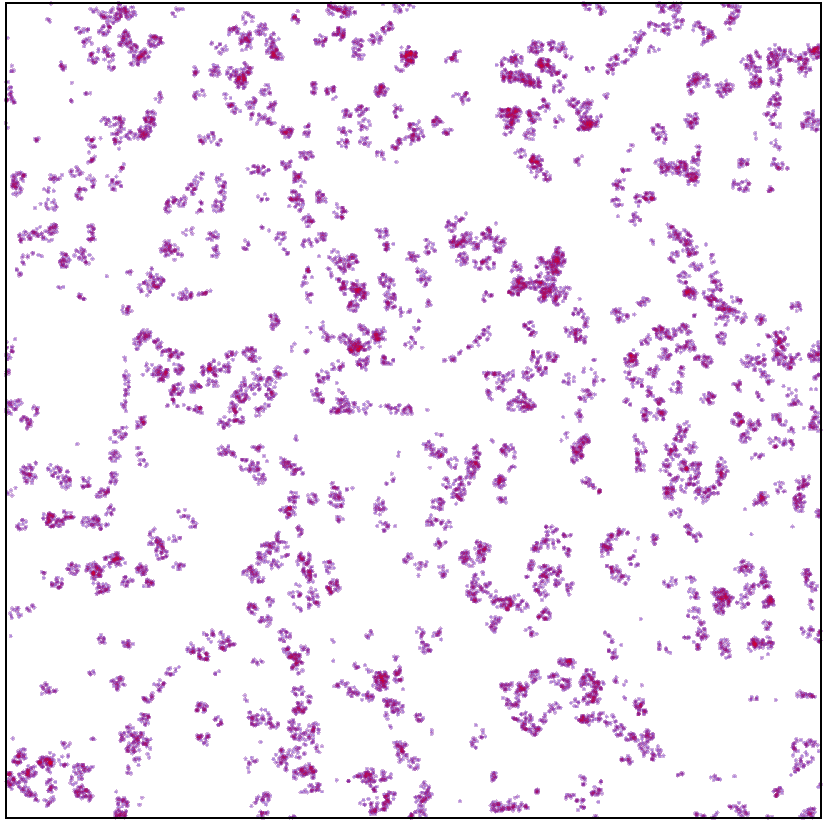}
    \caption{Samples of point processes of various geometries. The number of points ranges from 1000-13000.  
    }
    \label{Fig:exampels}
\end{figure*}

In this paper, 
we seek to generate realizations formed by a large number of particles, 
with both short and long range interactions. 
Figure \ref{Fig:exampels} shows some examples of distributions that we shall consider.
Currently, for such complex and diverse geometries,
which naturally appear e.g. in cosmology or turbulent flows in physics and atmosphere science, no model has been proposed in the literature on point processes. 
To address this problem, we shall introduce a statistical model developed from the maximum-entropy principle \citep{PhysRev.106.620}, 
to approximate such point process distributions and simulate new realizations.

Maximum-entropy models are based on the description of the distribution with a set of moments. Intuitively, this means that the model is 'as random as possible' under certain constraints, based on the information captured by the moments. There are three underlying problems in defining such models:
\begin{enumerate}
    \item Choosing the moments that will describe the distribution. They should be informative enough to capture the geometric structures characterizing the distribution. 
    On the other hand, they should be accurately estimated from a single observation, so the number of moments should not be too large.
    \item Specifying a model deriving from these moments. This can be done by defining a maximum entropy model such as the macro-canonical model (maximizing the entropy under expectation constraints), or the micro-canonical model (maximizing the entropy under path-wise constraints). 
    
    \item Generating new samples from the model. In the micro-canonical setup, 
    this can be done by minimizing an energy, that defines the set of admissible realizations. 
    The minimization method must make it possible to generate diverse low energy samples without being too costly in terms of calculation. 
\end{enumerate}

In this paper, we shall place our model in the micro-canonical setup, detailed in Section \ref{s.PP}. The main challenges reside in the problems 1 and 3.  
In this regard, we present 
multi-scale moments, new in the literature on point process, as well as a fast sampling algorithm based on gradient descent.

In Section \ref{s.GenerativeModel}, we present our method to address the problem of generating new samples: 
we minimize the energy of a new sample by moving the particles of an initial random configuration 
using the gradient of its energy with respect to the particles positions. 
In the point process literature, a classical method \citep{Tscheschel2006} consists in updating an initial random configuration by
successively replacing the particles one by one, with new particles located at random positions (we shall call this method random search in this paper). The major drawback of this method is its computational cost, as the optimization, which does not use gradient information to minimize the energy, requires a large number of energy evaluations.
In fact, this method has been applied to generate point processes formed by a few hundred particles. 
On the other hand, advanced methods in the modelling of textures and non-Gaussian stationary processes
allow for fast sampling by first drawing from an initial Gaussian distribution, and minimizing an energy by gradient descent on the amplitudes of the pixels of the image \citep{Portilla2000, gatys2015texture, bruna2018multiscale, zm-19}. 
Our approach leverages the efficiency of this sampling method, while ensuring that the resulting samples are atomic measures. 
The idea of moving the points according to their gradient is often  used in molecular dynamics \citep{PhysRevE.92.022119, PhysRevE.92.022120}, however it requires knowledge of the physical mechanisms behind the underlying process. Our statistical modeling approach has a potential to simulate new, complex particle configurations directly from one observation, when the underlying physical phenomena are very complicated to model.

This brings us to the other challenge that we address in this work: choosing the moments that we shall use to characterize the distribution.
In Section \ref{s.Wavelets}, we present the wavelet phase harmonic covariance moments for point processes.
These are spatial statistics based on coefficients computed from 
a wavelet transform of atomic measures, i.e. the convolution of the atomic measures with continuous local functions. 
It is known that the covariance between the wavelet coefficients capture only second-order correlations 
\citep[Section 5.2]{bremaud2013mathematical}, which are equivalent to the Bartlett spectrum \citep{10.2307/2334136}.
To capture information beyond second-order correlations, 
we apply a non-linear phase harmonic operator to the wavelet coefficients. 
This operator acts on the complex phase of the wavelet coefficients,
without changing their amplitude \citep{mzr-18}.
The covariance between the resulting coefficients allows one to capture particles interactions across different scales.
Compared to high-order correlation functions \citep[Section 12.4.2]{torquato}, our moments
have the potential to define a sufficient set of statistics,
while maintaining a small estimation error, 
which is similar to the second-order statistics.
Other statistics often used in the point process literature (e.g. the $k$ nearest neighbour distribution function suggested in \citet{Tscheschel2006}) have a number of elements that grows with the intensity.
Since there is only one observation, the number of moments should be limited, in order to control their estimation variance. 
The wavelet transform allows for direct control over the scales of the structures that we wish to capture, 
regardless of the intensity of the process. 
This property allows one to model point processes formed by a large number of particles with a limited number of moments.

The wavelet phase harmonic covariance descriptors are defined as spatial averages evaluated over a point process realization \citep{zm-19}. 
In practice, the calculation of such descriptors can be done by discretization of the observation window in the form of a grid of pixels.
However, making these descriptors differentiable with respect to the positions of the particles remains a challenge.
In this regard, we describe in Section~\ref{s.Scheme} a complete numerical scheme allowing one to solve this problem. It is based on a differentiable discretization of atomic measures. 
We further present a multi-scale optimization in the gradient descent, intended to avoid unwanted shallow minima of the energy.

In Section \ref{s.Numerical-Results}, we evaluate our model on some distributions exhibiting 
various geometric structures, like Cox point processes on the edges of
Poisson-Voronoi tessellations and on the Boolean model with circular
grains. 
Other processes we consider are Matern hard-core and cluster processes
driven by Poisson processes with turbulent intensities. 
Their intensities are sampled from a turbulent flow simulated from Navier-Stokes equations \citep{schneider2006coherent}. 
Besides the visual inspection of the samples from
our generative model, we evaluate second
order correlations
and compare the persistent homology diagrams,
that has been proven useful for topological data analysis (see
e.g. \citet{tda}). 

In Section \ref{s.comparison}, we numerically compare
our method with the classical approach developed in \citet{Tscheschel2006} in terms of the speed of simulation, 
and the quality and diversity of the syntheses. Besides using the 
evaluation methodologies in Section \ref{s.Numerical-Results}, we also 
use a statistical moment matching approach suggested in \citet[Chapter 6]{illian}. 
All the results can be reproduced by a software which is available at \url{https://github.com/abrochar/pp_syn}. 
A longer version of this paper is available, see \citet{brochard2020particle}.


\textbf{Notations}: For any integer $n\geq1$ and any $z \in \mathbb C^n$, we note $|z|$ the Euclidean norm of $z$, and $z^*$ is its complex conjugate. Let $\Cov(A,B)=\E[AB^*]-\E[A]E[B^*]$ denote the covariance between two complex random variables $A$ and $B$. 
Let $ \lb a , b \rb$ denote the Euclidean inner product between two vectors $a \in \Rdd$ and $b \in \Rdd$.

\section{Point process framework}
\label{s.PP}
In Section \ref{ss.gen-defs}, we define the elementary objects of point process theory and the notations that we will use in this paper. A more detailed introduction to point processes and stochastic geometry can be found e.g. in \citet{daley2007introduction, chiu2013stochastic}.
We then review, in Section \ref{ss.maxent}, the classical maximum-entropy models for point processes.

These models are theoretically well founded, but hard to sample from in general. Our model, presented in Section \ref{s.GenerativeModel}, takes inspiration from these, while being amenable to fast sampling.

\subsection{General definitions}
\label{ss.gen-defs}


Configurations of points (on the plane) are represented as  counting measures on  $(\mathbb R^2,\mathcal B)$, with $\mathcal B$ denoting the  natural Borel $\sigma$-algebra on $\mathbb R^2$. Recall that counting measures are locally finite  measures taking values in $\bar{\mathbb N} := \mathbb N \cup \{+\infty\}$. 
Let   $\Mi$  denote the space of all such measures on $(\mathbb R^2,\mathcal B)$, endowed with the $\sigma$-algebra
 $\mathcal M$ generated by the mappings $\mu \mapsto \mu(B)$, for $B \in \mathcal B$.
 For $\mu \in \Mi$, we will often use the following representation: 
\begin{equation}\label{e.mu-representation}
\mu = \sum_{1 \leq i \leq I}\delta_{x_i}, \  I \in\bar{\mathbb N},
\end{equation}
where $\delta_x$ is the Dirac measure having a unit atom at~$x$. 

Recall, a {\em push-forward} $F_\#\mu$ of a point measure~$\mu$ 
by a (measurable) function $F:\mathbb R^2\longrightarrow \mathbb R^2$ is simply the displacement of its atoms by the function $F$
$$F_\#\mu=\sum_i\delta_{F(x_i)}.$$
As a special case, for $x \in \mathbb R^2$, we define the translation $S_x\mu $ of $\mu$ by $x$, i.e. $S_x\mu(B) := \mu(B+x)$.

A counting measure $\mu\in\Mi$ is called simple if for all  $x \in \mathbb R^2$, $\mu(\{x\})=0$~ or~1 (in other words all atoms of $\mu$ in the representation~\eqref{e.mu-representation} are distinct). Simple counting measures can be identified  with their supports 
$\text{Supp}(\mu) := \{x \in \mathbb R^2 : \mu(\{x\})>0\}$ and in this regard we shall 
also write  $x \in \mu$ if $x$ is an atom of $\mu$, i.e., if $\mu(\{x\})>0$.

A point process $\Phi$ is a measurable mapping from an abstract probability space 
$(\Omega, \mathcal F, \mathbb P)$ to $(\Mi, \mathcal M)$. We will denote by $\mathcal L_\Phi$ the distribution of $\Phi$, that is the pushforward of the probability measure $\mathbb P$ by $\Phi$ on  $(\Mi, \mathcal M)$.
We say that a point process $\Phi$ is simple if $\mathbb P( \Phi \text{ is a simple measure})=1$. In this paper, for simplicity we shall only consider simple point processes.

Point process~$\Phi$ is called  {\em stationary} if its distribution $\mathcal L_\Phi$ is invariant with respect to all shifts $S_x$, $x\in\mathbb R^2$. It is said to be {\em ergodic} if 
the empirical averages (of real, measurable functions $f$ on $\mathbb M$, 
integrable with respect to $\mathcal L_\Phi$) 
over windows $W_s= [-s,s[\times[-s,s[$  increasing to $\mathbb R^2$ converge almost surely to the mathematical expectations 
\begin{equation}\label{e.Birkhoff}
\lim_{s\to\infty}\frac 1{|W_s|} \int_{W_s} f(S_x\Phi)\,dx=\mathbb E[f(\Phi)]=\int_{\mathbb M} f(\mu)\,\mathcal L_\Phi(d\mu),
\end{equation}
where $|W_s|$ stands for the Lebesgue measure of $W_s$, see~\citet[Chapter~12]{daley2007introduction}
for more details.

For a given $s>0$, we denote by $\M^s$ the set of counting measures on $W_s$, 
and $\mathcal M^s$ its induced $\sigma$-algebra. We will consider $W_s$ 
with addition and scalar multiplication modulo $W_s$.
Also we shall denote by $\bar S_x$ the corresponding shift operator on $\M^s$ with torus correction on the window $W_s$.

Let $\Phi$ be a point process on $\R^2$. One can only observe realizations of $\Phi$ on bounded subsets of $\R^2$. For the remainder of this paper, we shall consider realizations of point processes observed on a finite square window $W_s = [-s,s]^2$, for some $s>0$. We denote by $\bar\Phi$ the restriction of $\Phi$ to $W_s$, that is $\bP$ is a point process on $W_s$ such that, $\forall \, n \in \mathbb{N}, \forall \, (B_1, ..., B_n) \in \mathcal{B}(W_s)^n, \: (\Phi(B_1), ..., \Phi(B_n)) = (\bP(B_1), ..., \bP(B_n))$ in distribution (where $\mathcal{B}(W_s)$ stands for the Borel $\sigma$-algebra on $W_s$).
A realization of $\Phi$ observed on $W_s$ is therefore a realization of $\bP$, and will be noted $\bp$.

\subsection{Maximum entropy models for point processes}
\label{ss.maxent}

Maximum entropy models are based on the following intuitive idea: given an observation pattern, we aim at finding new patterns that are similar to, but different from the observation. To this end, we define a notion of similarity by choosing a set of statistics
that will be computed on the observation and on the new patterns. The two will be considered similar if their statistics match. Furthermore, if the chosen statistics describe sufficiently well the point process behind our observation, we do not want to add any more constraints, that is, we want to find new patterns 'as random as possible', under the constraints defined by the statistics. This can be formalized by maximizing the entropy of the model.

This section defines both macro-canonical 
and micro-canonical models for a point process $\bar\Phi$ observed in the square window $W_s$. These models rely on maximizing the 
entropy a probability distribution under a set of moment constraints. 
They are used in large classes of stochastic models \citep{Geman2}, and will inspire our particle gradient descent model.

\subsubsection{Point process entropy}
The notion of entropy is naturally defined only for random objects in discrete state spaces.
Even if a mixture of the  differential and discrete entropy can be considered  for point processes \citep{baccelli-entropy},
it is more natural to consider in this context the Kullback-Leibler (KL) divergence with respect to 
a reference distribution, naturally taken to be the homogeneous  Poisson point process distribution \citep{inbook}. More specifically, let us denote by $\mathcal{L}_0$ the Poisson  distribution on~$W_s$. We define the KL divergence  of a point process $\bar\Phi$ on $W_s$ with distribution $\mathcal{L}_\Phi$ (here, we replaced $\bP$ by $\Phi$ for notations simplicity),
\begin{equation}
    \text{KL}(\mathcal L_\Phi;\mathcal L_0) := \int_{\mathbb M^s}\frac{d\mathcal L_\Phi}{d \mathcal L_0}(\mu) \log\frac{d\mathcal L_\Phi}{d \mathcal L_0}(\mu)\,\mathcal{L}_{0}(d\mu),
\end{equation}
provided $\mathcal L_\Phi$ is absolutely continuous w.r.t. $\mathcal L_0$, denoting by $\frac{d\mathcal L_\Phi}{d \mathcal L_0}$
the corresponding density (otherwise $\text{KL}$ is set to~$\infty$). 

\subsubsection{Maximum entropy models}\label{sss.maxent}
With the KL divergence as a notion of entropy for point processes, we can now define the macro-canonical and micro-canonical models. These models are distributions of maximum entropy under different types of constraints. When considering these models as approximations of a point process $\bP$, the constraints are usually built as functions of the distribution of $\bP$, or functions of samples from $\bP$. 
Consider a mapping $K : \M^s \longrightarrow \mathbb C^d$, for some $d<\infty$ (one can think of, for instance, estimators of the $k$ nearest neighbours distribution functions, $D_k(r)$, such as in \citet{Tscheschel2006}).

\paragraph{Macro-canonical model} The macro-canonical model is defined as the distribution $\mathcal{L}$ of a point process $\Xi$ on $\mathbb M^s$ that  {\em minimizes}  the  KL divergence $\text{KL}(\mathcal L,\mathcal L_0)$ under {\em expectation constraints}: $\mathbb E(K(\Xi))=a$, for some vector of constraints, e.g.  $a=\E( K(\bP))$ or $a = K(\bp) $. 
Under some technical assumptions (in particular, having density
with respect to the reference Poisson distribution), the solution of the macro-canonical model is given by the Gibbs point process \citep[Section 1.3]{inbook}.
Sampling from the macro-canonical model is usually computationally very expensive \citep{bruna2018multiscale}. Therefore, we shall focus on the micro-canonical model, defined in the following.

\paragraph{Micro-canonical model}\label{microcan}

The micro-canonical model is defined by replacing the expectation constraints 
$\mathbb E(K(\Xi))=a$ with pathwise constraints. 
Let $\bp\in \M^s$ be our observation sample, of unknown distribution. For all $\mu \in \M^s$, we define the energy of $\mu$ as:
\begin{equation}
    \label{energy}
    E_{\bp}(\mu) := \frac{1}{2}|K(\mu) - K(\bp)|^2.
\end{equation}

The micro-canonical set of level $\epsilon$, for some $\epsilon>0$, is defined as
\begin{equation}
\label{e.om-eps}
    \om := \{ \mu \in \M^s \, : \, E_{\bp}(\mu) \leq \epsilon\}.
\end{equation}
The micro-canonical  model is defined as the distribution
$\mathcal L$ that minimizes the KL divergence 
with respect to the reference distribution $\mathcal L_0$  under {\em pathwise constraints} requiring $\mathcal L$ to be supported on~$\om$:  
\begin{align}\label{e.micro}
\arg\min_{\mathcal L}\quad &\text{KL}(\mathcal L,\mathcal L_0) \\
\text{given}\quad & \int_{\M^s} \mathbbm1(\mu\in \om) \mathcal L(d\mu)=1,\label{e.micro-constraint}
\end{align}
where $\mathbbm1(\cdot)$ is the indicator function.
If $\mathcal{L}_0(\om)>0$, the solution to this problem~\eqref{e.micro} \eqref{e.micro-constraint}, is the measure $\mathcal{L}$ having a uniform density on $\M_s$ given by $\frac{d\mathcal L}{d \mathcal L}_0(\mu) = \frac{1}{\mathcal{L}_0(\om)}\mathbbm1(\mu\in \om), \: \mathcal{L}_0-a.s.$

In order to consider the micro-canonical model as a good approximation of the observation distribution, one usually aims at finding $K$ satisfying the following properties:

\begin{itemize}
\item[(P1)] \textit{Concentration property}: The value of $K(\bar\Phi)$ should concentrate around its mean, i.e. $K(\bar\Phi)\simeq \mathbb E[K(\bar\Phi)]$ with high probability. A natural assumption is that the variance of $K(\bar\Phi)$ is small. 
\item[(P2)] \textit{Sufficiency property}: The moments $\mathbb E(K(\bP))$, should {\em characterize the unknown  distribution} as completely as possible. It requires that $K$ has a strong (distributional) discriminate power.
\end{itemize}

A natural framework allowing one to address (P1) and (P2) is by defining the descriptors $K=(K_1,\ldots,K_d)$ as a vector of empirical averages 
\begin{equation}\label{e.K-average}
    K_i(\mu)=\frac 1{|W_s|} \int_{W_s} f_i(\bar S_x\mu)\,dx \qquad \mu\in\M^s,    
\end{equation}
for a sufficiently rich class of functions
$f_i$ on $\M^s$, and relying on the ergodic assumption~\eqref{e.Birkhoff} regarding $\Phi$.

These properties are needed in order to have a model that reproduces typical geometric structures in $\Phi$, and generates diverse samples.

In this paper, we shall consider that the number of points of our model in $W_s$ is fixed. In such a case, it is customary 
to take the homogeneous Poisson point process distribution conditioned on having exactly $n$ points in $W_s$ as the reference measure,
which is equivalent to $n$ points sampled uniformly, independently in~$W_s$. We will note this distribution $\mathcal{L}_0^n$.

Sampling from the uniform density on $\om$ efficiently remains an open problem. 
In the literature on stochastic process modelling, 
most sampling algorithms rely on the following method: 
one first samples from an initial, high-entropy measure, and iteratively minimize the energy (cf. \eqref{energy}) of this sample until it reaches $\om$.
By choosing a high entropy initial measure, one hopes that the resulting model also has a high entropy. Recall that the micro-canonical model has the highest entropy supported on $\om$. 
Contrary to the classical methods in the point process literature \citep{Tscheschel2006, ppsim}, which relies on random search,
popular methods in image modelling use gradient descent to perform fast sampling in the micro-canonical set. 
However, optimizing the values of the image pixels 
does not guarantee that the resulting sample is an atomic measure.
For these reasons, in what follows we propose a model based on the transport of a Poisson point process via a gradient descent algorithm.

\section{Particle gradient descent model}
\label{s.GenerativeModel}

In Section~\ref{ss.Gradient-descent}, we introduce the particle gradient descent model, that uses gradient descent on the positions of the particles of the sample. This model consists in using the gradient of a prescribed energy to move the points of an initial random configuration, until we obtain a pattern similar (in an informal sense) to the observation. We then present in Section~\ref{sss.Symmetries} a theorem stating that this model preserves some basic invariances of the original distribution. This result, allowing us to gain some understanding about the entropy of our model, extends the results of \citet{bruna2018multiscale}. In Appendix \ref{s.relax}, we present some ideas about how to relax the hypotheses made in this paper, in order to build a model better suited for real world data.


\subsection{Particle gradient descent model}
\label{ss.Gradient-descent}

As in Section~\ref{ss.maxent}, let $\bp \in \M^s$ be our observation sample of unknown distribution, and $K : \M^s \longrightarrow \mathbb C^d$, for some $d<\infty$ a mapping defining our descriptors. We note the resulting energy $E_{\bp}$ (cf. \eqref{energy}).

Let $\bp_0$ sample from an initial distribution that we choose as $\mathcal{L}_0^{\bp(W_s)}$ (i.e. the same number of particles as $\bp$, drawn uniformly and i.i.d.). We minimize the energy of $\bp_0$ through its gradient with respect to the particles positions. More precisely, we define the mapping 
\begin{equation}
\label{e.particle-gradient}
\begin{array}{ccccc}
F& : & \M^s & \longrightarrow & \M^s\\
 & & \mu=\sum_i \delta_{x_i} & \longmapsto & \displaystyle{\sum_i \delta_{x_i - \gamma\nabla_{x_{i}}E_{\bp}(\mu)}} \\
\end{array}
\end{equation}
for some gradient step $\gamma>0$. The measure $F(\mu)$ can be seen as  the push-forward  $F_{\mu\#}\mu$
of the measure $\mu$
by the   mapping $F_{\mu}(x) := x - \gamma\nabla_{x}E_{\bp}(\mu)$ (see e.g. \citet{molchanov2002steepest} for more details about steepest descent methods on spaces of measures).
Note that the function $F_\mu$ depends on the measure $\mu$ which is pushed forward. For  any initial point measure  $\bar\phi_0\in\M^s$ 
we define   the successive  point measures:
\begin{equation}\label{e.pushforward-iterations}
\bar\phi_n:=F_{\bar\phi_{n-1}\#}\bar\phi_{n-1},\qquad n\geq 1.
\end{equation}

\paragraph{Pushforward of the point process distributions}
The pushforward operation $F_{\#\mu}$
on $\M^s$ induces the  corresponding pushforward operation on the probability measures on $\M^s$, which are distributions of point processes. We denote this latter by $\mathcal{F}_{\#}$: For a probability law $\mathcal L$ on  $\M^s$
$\mathcal{F}_{\#}\mathcal L(\Gamma) := \mathcal L(\{\mu\in \M^s: F_{\mu\#}\mu\in\Gamma\})$, for any $\Gamma \in \mathcal M^s$. Then, for an initial probability law $\mathcal L_{\bp_0}$ on $\M^s$
we define   the successive  probability laws 
\begin{equation}\label{e.pushforward-iterations-law}
\mathcal L_{\bP_n}:=\mathcal{F}_{\#}\mathcal L_{\bP_{n-1}},\qquad n\geq 1.
\end{equation}
Note that $\mathcal L_{\bP_n}$
is the distribution of the point process $\bar\Phi_n$ obtained by $n$ iterations of~\eqref{e.pushforward-iterations} starting from $\bP_0$ having law $\mathcal L_{\bP_0}=\mathcal L_{0}^{\bP(W_s)}$. Our model is defined by setting a fixed number of iterations as a stopping rule.

Observe that our model takes inspiration from the micro-canonical model, however there is no guarantee that the optimization reaches $\Omega_\epsilon$ (defined in \eqref{e.om-eps}), for any $\epsilon>0$. By setting a fixed number of iterations and not rejecting any configuration, we make the implicit assumption that our model reaches a low energy level.
In practice, one
can use classical line-search methods in the optimization to adjust the $\gamma$ in \eqref{e.particle-gradient}, 
so as to ensure that the energy decreases as $n$ grows.

\subsection{Leveraging invariances} \label{sss.Symmetries}

One can leverage some a priori known invariance properties of $\Phi$ (for instance stationarity or isotropy), by building a model that satisfies the same invariance properties as $\Phi$. 
By using the descriptor $K$ with the same invariance, 
the particle gradient model respects these invariance properties. 
In particular, we obtain a stationary point process model 
when $K$ is defined by the empirical averaging \eqref{e.K-average}.

This requires some explanation, since 
invariance properties of the distribution of $\Phi$ 
do not, in general, imply any natural invariance of its restriction $\bar\Phi$ to $W_s$.
Indeed, while some invariances can  be observed on the torus for the distribution of $\Phi$ on $\mathbb R^2$ (the most popular being translation invariance), it does not imply the same for $\bar\Phi$ with respect to the translation on ~$W_s$. The latter, called in this paper {\em circular stationarity}, 
requires also  $\Phi$ to be periodic.
However, circular stationarity of the generated point process on large window $W_s$ (as a distributional  approximation of $\bar\Phi$) can be considered as a desirable ersatz of the stationarity of $\Phi$.
Indeed, in what follows we shall formulate a result saying that, when $K$ and  the distribution of  $\bP_0$ are invariant with respect to  some subset of {\em rigid circular transformations} on $W_s$, then the resulting model satisfies this property as well.

More specifically, a {\em rigid circluar transformation} on $W_s$
is an invertible operator $T$  on $W_s$ of the form $Tx := Ax + x_{0}$ for some orthogonal matrix $A$ with entries in $\{-1,0,1\}$ and $x_0\in W_s$.
Note that the matrix A is restricted in integer entries for $T$ to be a well defined invertible operator. It encapsulates translations, flips, and orthogonal rotations.

We say that: 
\begin{itemize}
    \item The initial probability law  $\mathcal L_{\bar\Phi_{0}}$ of the model is invariant to the action of $T$ if $\forall \: \Gamma \in \mathcal \M^s, \: \mathcal L_{\bP_0}(T_\#^{-1}(\Gamma)) = \mathcal L_{\bP_0}(\Gamma).$
    
    \item The descriptor $K$ is invariant to the action of $T$ if  $\forall \: \mu \in \M^s, \: K(T_\#\mu) = K(\mu)$.
\end{itemize}

\begin{theorem}
\label{th.Invariance}
 Let $T$ be a rigid circular transformation. Let  $\bar\Phi_0$ be a point process on $W_s$ such that its distribution $\mathcal L_{\bar\Phi_0}$ is invariant to the action of $T$ and  let $K$ be a  descriptor invariant to the action of $T$.  Then, for all $n \in \mathbb N, \: \mathcal L_{\bar\Phi_n}$ defined as the push-forward of $\mathcal L_{\bar\Phi_0}$ by~\eqref{e.pushforward-iterations-law} is invariant to the action of $T$.
\end{theorem}

A proof of the above result is given  in Appendix~\ref{s.Invriance-proof}.
This property can guarantee distributional symmetries in our model with respect to the original  distribution, not stated in the classical approach of \citet{Tscheschel2006}.
The result itself is inspired from \citet{bruna2018multiscale}, where the preservation of invariance is proven for the gradient descent model in the pixel domain.
Observe, the  invariance of the distribution of the point process $\bar\Phi_n$ increases the diversity of the generative model samples. 
Our descriptor $K$ proposed in Section~\ref{ss.Covariance} will be 
invariant with respect to all circular translations.
This will be achieved by computing statistics of $\bar\Phi$ in the form of spatial averages~\eqref{e.K-average} with periodic boundary condition. This boundary condition means the use of the shift operator $\bar S_x$ in ~\eqref{e.K-average}, which can be interpreted as a torus correction on $W_s$. 


A drawback is that such a boundary condition introduces a statistical bias to the  spatial average~\eqref{e.K-average} as an estimator of $\mathbb E[K(\bP)]$ in the case of a non periodic $\Phi$ over $W_s$. 
One can expect, however, that when the window size is large enough and spatial correlations of the patterns are not too large, this  border effect becomes negligible.

\section{Wavelet phase harmonic descriptors}

\label{s.Wavelets}

In this section we present a family of descriptors that we will use, in conjunction with the particle gradient descent model, to capture and reproduce complex geometries of point processes. 

Classical descriptors for spatial point process usually include statistics more or less directly related to the pair correlation function, such as Ripley's $K$-function, Besag's $L$-function, or the radial distribution function \citep[Section 4.5]{chiu2013stochastic}. All of these functions only capture second order correlations of the process.
Other usual functions are the empty space function or the $k$-nearest neighbors function (see~\citet[Section 2.3.4 and 4.1.7]{chiu2013stochastic}).
In \citet{Tscheschel2006}, the authors advocate the use of the $k$-nearest neighbors distribution function, with a $k$ significantly greater than $1$. In addition to being non-differentiable, these moments suffer from another drawback. If one wants to capture geometric structures formed by the particles, up to a fixed scale, the number of moments (i.e. the $k$ nearest neighbours) will grow linearly with the number of particles forming such structures. This can become a problem if the intensity of the process is large, both computationally, and from a statistical point of view, as the variance of the moments may become large when estimated from a single observation.

For this reason, we choose in this paper to use descriptors for which the spatial range of structure captured is independent of the intensity of the process, and the computational time is linear in the number of points. As a result, this method would become much faster for large samples, as the number of statistics would remain constant. These descriptors, built 
upon the wavelet transform of a random configuration,
are adapted 
from \citet{zm-19}. 
They have shown high quality results in modelling geometric structures in texture images and turbulent flows.

We begin, in Section \ref{wph}, by presenting wavelet transform for counting measures, and their so called phase harmonics, which are derived from complex wavelet coefficients by applying a multiplication operator on their phase. In Section \ref{ss.Covariance}, we explain how wavelet phase harmonics can be used to capture dependencies between the wavelet coefficients of counting measures,
and detail the choice of the descriptors that we use for numerical experiments. 

\subsection{Wavelet transforms and their phase harmonics}\label{wph}

Informally, a wavelet $\psi : \R^2 \mapsto \mathbb C$ is a function that is localized both in the space and the frequency domains. Convoluted with an input signal, they allow one to capture its local geometric structure, at a given scale (see e.g. \citet[Section 7]{mallatbook}). To capture information at different scales in the signal, we build a family of wavelets by rotations and dilations of the wavelet $\psi$.
They constitute the foundation of the descriptors that we propose to use, in conjunction with our generative model described in Section~\ref{ss.Gradient-descent}. 

\subsubsection{Wavelet transform}\label{w_transf} 
The wavelet transform is a powerful tool in image processing to analyze 
signals presenting local geometric structures of different scales. 
Oriented wavelets have already been considered, e.g. to analyze anisotropy properties of planar point processes (see e.g. \citet{RAJALA2018141}).
We shall also use oriented wavelets which allow to capture edge-like geometric structures in the observation. 
Specifically, we choose bump steerable wavelets introduced in \citet{mzr-18}. 
They are defined by the translations, dilations and rotations of a complex analytic function $\psi(x) \in \C$ with $\int \psi(x)\, dx = 0$ and $\int |\psi(x)|\, dx < \infty$.

In what follows we first define the wavelet transform for the counting measures in $\M^s$ which are constructed from the function $ \psi$.
Let us denote the Fourier transform of $ \psi$ for $\omf \in \Rdd$ by $\what \psi(\omf) = \int  \psi(x) e^{-i \lb \omf , x \rb } dx$.
By construction, the function $\psi$ is centered at a frequency $\wxi \in \R^2$, and it has a compact support in the frequency domain, as well as a fast spatial decay. Assume that $|\psi(x)|$ is negligible if $|x| > C$, for some $C>0$. 

Let $r_{\theta}$ denote the rotation by angle $\theta$ in $\Rdd$. 
Multiscale steerable wavelets are derived from $\psi$ with dilations
by factors $2^j$ for $j \in \Z$, and rotations $r_\theta$ over
angles $\theta = 2 \ell \pi / L$ for $0 \leq \ell < L$, 
where $L$ is the number of angles between $[0,2\pi)$. 
The wavelet at scale $j$ and angle $\theta$ is indexed by its central frequency
$\la := 2^{-j} r_{-\theta}\, \wxi\in \Rdd$,
and it is defined by
\begin{equation*}
\label{steerable}
\psi_{\la} (x) =  2^{-2 j} \psi (2^{-j} r_{  \theta  } x ) ~~\Rightarrow~~
\what \psi_{\la} (\omf) = \what \psi (2^{j} r_{ \theta   }  \omf ) .
\end{equation*}
Since $\what \psi(\omf)$ is centered around $\wxi$, it results that $\what \psi_\la(\omf)$ is centered around the frequency $\la$. The wavelet $\psi_\la$ at scale $j$ has negligible amplitude for $|x|> 2^j C$.


For a counting measure $\mu \in \mathbb M^s$,
we typically consider only the wavelets having spatial support \footnote{More precisely, where the wavelet norm is non negligible} contained in $\W$ by limiting the scale $j < \jmax$ such that $2^\jmax C \leq 2s$.
Scales equal or larger than $\jmax$ are carried by a low-pass filter whose frequency support is centered at $\la = 0$. It is denoted by $\psi_{0}$.
Let $\La$ be a frequency-space index set including
$\la  = 2^{-j} r_{-\theta} \wxi$ for 
$ 0 \leq j < \jmax$, $0 \leq \ell < L$, 
and $\la= 0$. As we eliminate $j < 0$ in  $\La$ to ignore structures smaller than $C$ in $\W$, the parameter $\xi_0$ will be adjusted in Section \ref{ss.pix} for a suitable choice of $C$.

The {\em wavelet transform} of a counting measure 
$\mu\in\M^s$ observed in the finite window~$W_s$, 
is a family of functions 
obtained by the convolution of $\mu$ with 
periodic wavelets $\psi_{\la}^s$, 
\begin{equation}\label{periodic}
    \mu \star \psi_\la(x) = 
    \int_{\W} \psi_{\la}^s (x-y) \mu(dy) , \quad \la \in \La . 
\end{equation}
They are defined with {\em periodic edge connection}, i.e. 
at $x = (x_1,x_2) \in \W, \: 
\psi_{\la}^s (x_1,x_2) := \sum_{n_1,n_2 \in \Z} \psi_\la(x_1 + 2s n_1, x_2 + 2s n_2)$. 
The integral \eqref{periodic} can be interpreted as a shot-noise, which is thus well-defined because $\int_{\R^2}|\psi_\lambda(x)|dx<\infty$. 
We denote the wavelet coefficients of $\mu\in\M^s$ by $\displaystyle{ \{  \mu \star  \psi_{\la}(x) \}_{\la \in \La, x \in \W}}$.

\textbf{Remark:} As the wavelet transform is a linear transformation of a counting measure, it is known that the covariance between $\bP \conv \psi_\la (x)$ and $ \bP \conv \psi_{\la'} (x')$ depends only on the mean intensity and second-order correlations of a stationary point process $\Phi$ \citep[Eq.~(5.27)]{bremaud2013mathematical}, which only gives partial information on the process distribution.

\subsubsection{Wavelet phase harmonics}\label{wph-def}

To capture random geometric structures of different scales occurring simultaneously (i.e. at nearby $x$ and $x'$) in a given process, one can compute the covariance of the wavelet coefficients at different scales. 
However, due to the frequency localization property of the wavelets, such covariance can be close to 0, even though the wavelet coefficients are not independent. To capture such dependencies, one can use a non linear operator on the transforms, to superimpose their frequency support. This section, along with the following, details this non linear operator, and the resulting covariance moments.

{\em Phase harmonics} \citep{mzr-18} of a complex number $z \in  \mathbb C$ are defined 
by multiplying its phase $\varphi(z)$ by integers $k$, while keeping the modulus constant, i.e. 
\[
 \forall \: k \in \mathbb Z, \: [z]^k := |z|e^{ik\varphi(z)}. 
\]
Note that $[z]^0=|z|$, $[z]^1=z$, and $[z]^{-1}=z^\ast$ (complex conjugate of~$z$). More generally,  $([z]^k)^\ast = [z]^{-k}$ and $| [z]^k | = |z|$ for $k \in \Z$. 

We apply the phase harmonics to adjust the phase of the wavelet coefficients. 
For all $ x \in W_s^2, \la \in \La$, $k \in \Z$, let's denote the {\em wavelet phase harmonics} of $\phi \in \M^s$ by 
\[
    [\phi \conv \psi_\la (x)]^k = |  \phi \conv \psi_\la (x) |  e^{i k \varphi( \phi \conv \psi_\la (x) ) }   . 
\]
The phase of the wavelet coefficient $\varphi( \phi \conv \psi_\lambda(x) )$ is multiplied by $k$, whereas the modulus $| \phi \conv \psi_\lambda(x) | $ remains the same for all $k$. 
Note that the wavelet phase harmonics at $k=1$ are exactly the wavelet coefficients .

As illustrated in \citet{zm-19}, when $\phi$ is a realization of a stationary process, the frequency support of $\phi \conv \psi_\lambda$, which is centered around $\lambda$, is shifted and dilated by the phase harmonics. As a consequence, $[\phi \conv \psi_\lambda]^k$ has a frequency support roughly centered around $k \lambda$. 
This non-linear frequency transposition property is crucial to capture dependencies of the wavelet coefficients across scales and angles, as we shall detail next.

\subsection{Wavelet phase harmonic covariance descriptors}
\label{ss.Covariance}


A classical way to capture dependencies between wavelet coefficients is to compute their higher order moments. However, as the order grows, so does the variance of the moment estimator (which may violate (P1)).
Based on the frequency transposition property of the phase harmonics (see Section \ref{wph-def}), 
we shall explain how to capture dependencies between the wavelet coefficients at different locations and frequencies by computing the covariance between wavelet phase harmonics. 
Note that the wavelet phase harmonics do not increase the amplitude of the wavelet coefficients with $k>1$. This approach may thus significantly reduce the variance of the descriptor $K$ (to satisfy (P1)) compared to the higher order correlations, while still capturing information beyond second-order correlations (to satisfy (P2)). %

The wavelet phase harmonic covariance of $\bP$ is defined by 
\begin{equation}
    \Cov ( [ \bP \conv \psi_\la (x) ]^k,  [ \bP \conv \psi_{\la'} (x') ]^{k'} )  , 
    \label{eq:whpcov}
\end{equation}
for pairs of $(x,x') \in W_s \times W_s$, $(\la,\la') \in \La^2$, and $(k,k') \in \Z^2$.

In particular when $k \neq 1$ or $k' \neq 1$, the covariance measures the dependencies between the wavelet coefficients.
As explained in \citet{mzr-18, zm-19}, for a stationary process $\Phi$, 
the overlap between the frequency support of $[\bP \conv \psi_{\lambda}]^k$ and that of $[\bP \conv \psi_{\lambda'}]^{k'}$ is necessary for the wavelet phase harmonic covariance to be large.  Due to the frequency transposition property of the wavelet phase harmonics, 
it is empirically verified that the covariance at $k \la \approx k' \la' $ is often non-negligible when the process is non-Gaussian (i.e. has structures beyond second order correlations). We shall also follow this empirical rule to select a covariance set $\Gaw$ (specified in detail in Section \ref{gamma_choice}) to describe point processes.

Let $v_{\lambda,k} = \E( [\bP \conv \psi_\la (x)]^k )$. We define the descriptors $K(\mu)$ using~\eqref{e.K-average}, as empirical estimators of moments.
Additionally, let us denote $\mu_{\la,k}(x):=[ \mu \conv \psi_\la (x) ]^k -  v_{\la,k}$.
Taking the spatial average~\eqref{e.K-average} gives the descriptor of the form: 

\begin{equation}\label{eq:wphK}
    K(\mu) = \biggl( \frac{1}{|W_s|}\int_{W_s} 
   \mu_{\la,k}(x) \: \mu_{\la',k'}(x-\tau')^\ast 
        dx \bigg)_{(\la,k,\la',k', \tau') \in \Gaw}.
\end{equation}
As $\bP$ is circular-stationary, \eqref{eq:whpcov} depends only on $x-x'$, it suffices to use the vectors $\tau'$ to measure the differences between $x$ and $x'$. Note also that $K(\mu)$  is invariant with respect to any circular translation $\bar S_x$ of $\mu\in\M^s$ on  $x\in W_s$.

In the numerical computation, we shall replace $v_{\la,k}$ in \eqref{eq:wphK} by $\bar{v}_{\la,k} = \frac{1}{|\W|} \int_{ \W } [ \bp \conv \psi_\la (x) ]^k dx$ as a plug-in estimator for the first-order moment  $v_{\lambda,k}$. The $K(\bp)$ modified in this way becomes an empirical estimator of the covariances in \eqref{eq:whpcov}. This is a good approximation of $K(\mu)$ as the estimation variance of the covariance moments is typically much larger than that of the first-order moments. 

\subsection{Choice of the covariance set $\Gaw$ in \eqref{eq:wphK}}
\label{gamma_choice}

Rather than detailing the full list of elements in $\Gaw$, we provide an intuitive way to choose the set $\Gaw$. For the full list, see \citet[Section 4]{brochard2020particle}. Overall, the total number of elements in $\Gaw$ is in the order of $O(L^2 J^2)$.
Note that the smallest structures that the descriptors can capture depend on the spatial support C of the wavelet $\psi$. Information about structures smaller than C can be added in a post-processing step will be explained in Section 5.  
 
\begin{itemize}
    \item Choice of $J$: The covariance set $\Gaw$ depends on the parameter $J$, which is the maximal scale of the wavelet transform. A suitable choice for this parameter $J$ would be one allowing for a good trade-off between satisfying the sufficiency of $K$, while maintaining the concentration property (cf. properties (P1) and (P2) from Section \ref{ss.maxent}). 
    \item The parameter $\tau'$ is chosen so that each wavelet is translated in a particular direction in order to capture correlations along nearby edges in the observation.
    \item Choice of $(\lambda, \lambda', k, k')$: These parameters are chosen in order to capture 2nd-order correlations, as well as dependencies between wavelet coefficients at different scales and orientations, both with and without phase information, based on a rule of thumb that $k \la \approx k' \la'$ due to the frequency transposition property of the wavelet phase harmonics. Figure \ref{fig:csrwph} shows the impact of using 
    phase harmonics coefficients (with $k,k' \not= 1$) compared to 2nd-order correlations (only $k=1,k'=1$).
    We see that both syntheses deviate from complete spatial randomness, but important structures, such as vortexes, are better reproduced when incorporating the non-linear coefficients.
    

\end{itemize}
{\setlength{\tabcolsep}{1.5pt}
\begin{figure}[h!]
    \centering
    \begin{tabular}{ccc}
        \includegraphics[scale=.4]{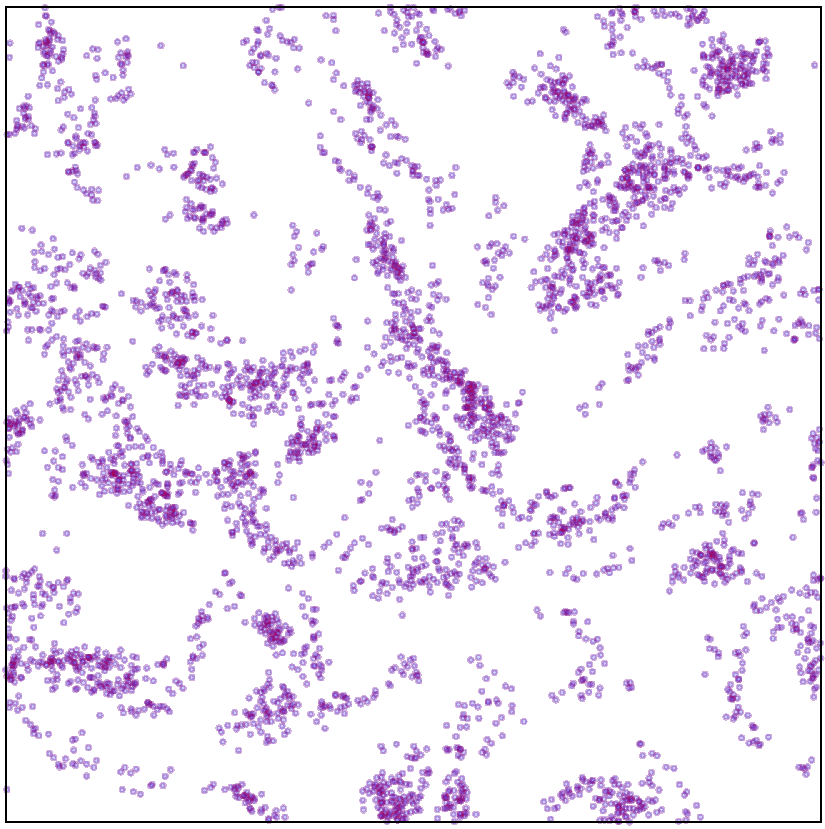}&
        \includegraphics[scale=.4]{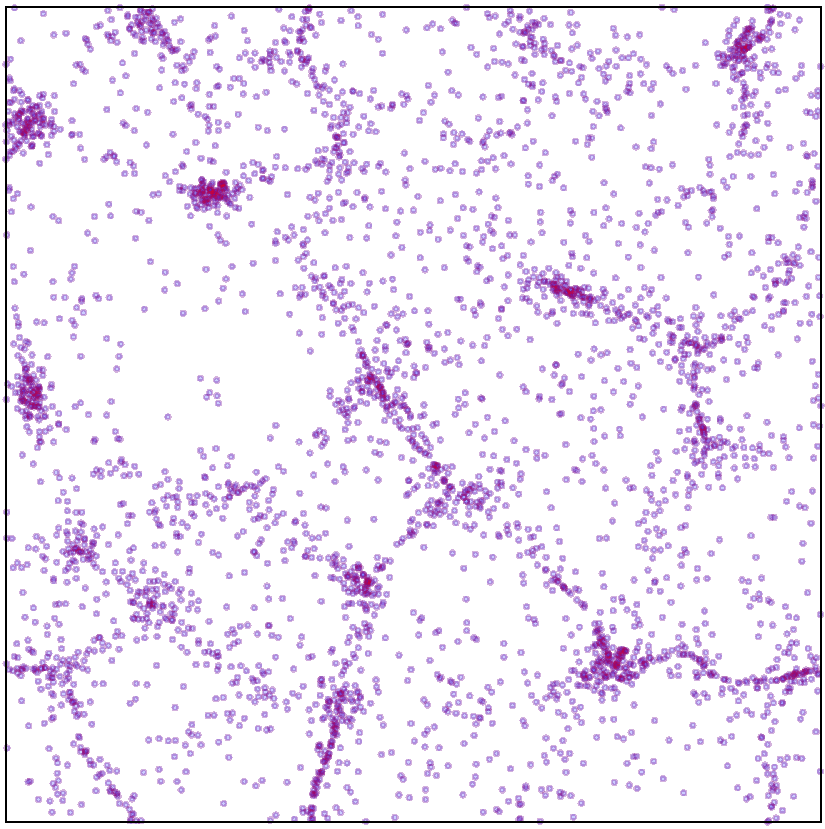}&
        \includegraphics[scale=.4]{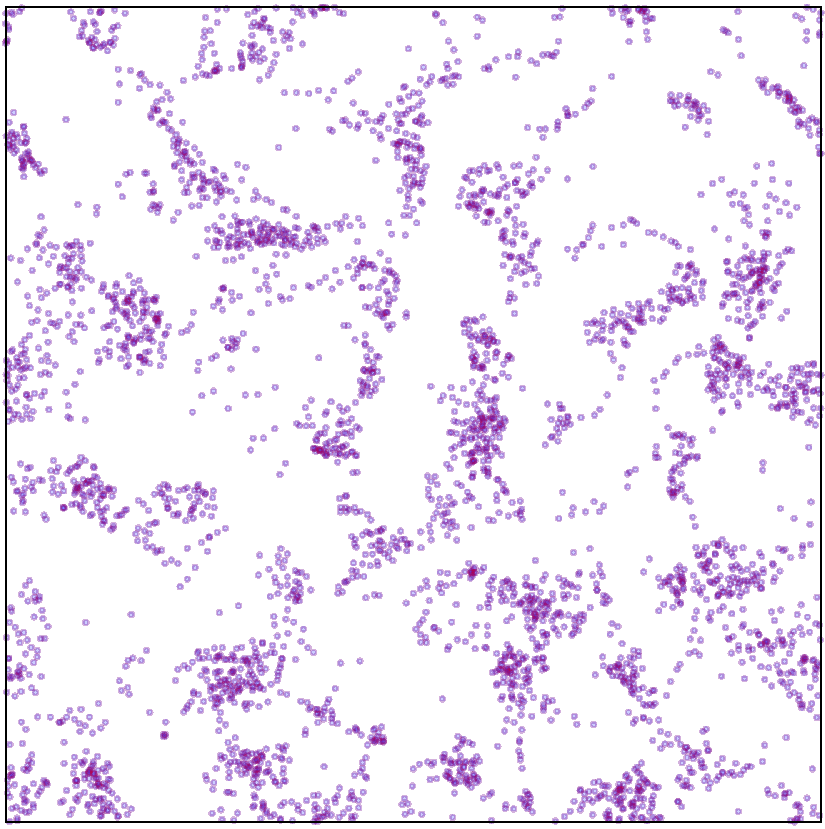}
    \end{tabular}
    \caption{
    Left: Observation of a turbulence Poisson process (original), middle: synthesis with wavelet covariances without phase harmonics non-linearity 
    (i.e. only $k=k’=1$ in $\Gaw$ in \eqref{eq:wphK}), right: synthesis with wavelet phase harmonic covariance descriptors (full $\Gaw$)}
    \label{fig:csrwph}
\end{figure}}

\section{Numerical scheme for particle gradient descent} 

Calculating the wavelet phase harmonic covariances can be computationally demanding (due to the calculation of two integrals). In order to gain some efficiency, we can perform the computations in a discrete domain. However, the energy needs to remain differentiable with respect to the positions of the points in the pattern. We propose a method, consisting of a Gaussian smoothing of the configuration of points, to address this problem. Building on that method, we then present two technical aspects of the sampling method.

\label{s.Scheme}
In this section, we discuss a complete numerical scheme to generate samples from
the particle gradient descent model, defined
with the wavelet phase harmonic descriptors presented in Section~\ref{s.Wavelets}.  
It is composed of the following ideas:    
\begin{itemize}
    \item {\em Discretization} for an approximate calculation of the covariance of the wavelet phase harmonics: necessary to accelerate the calculation of the descriptor and the gradients. 
    \item {\em Multiscale optimization}: allowing one to avoid shallow local minima in the gradient descent model. At each scale, we use a quasi-Newton gradient-descent method for greater efficiency.
    \item Final  {\em blurring} (optional): to add a priori information on structures whose size is smaller than $C$ into the model samples. It helps to get rid of some clusterisation (clumping) artifact caused by the initial discretization. 
\end{itemize}

\subsection{Discretization}
\label{ss.pix}

\subsubsection{Differentiable discretization of atomic measures}

To compute the descriptor $K$ in~\eqref{eq:wphK} for a point measure $\mu$, we need to integrate functions over the observation window $W_s$ (first for the convolution operators, then for the averages). Computationally efficient integration requires discretization of the atomic measure.
The main difficulty is to do it in such a way that the (periodic) convolutions of the discretized  atomic measures with wavelets, as in \eqref{periodic}, remain differentiable
with respect to the positions of the original atoms in $\mu$, so that we can still perform gradient descent. Classical finite element methods may not achieve this goal efficiently.

\begin{figure}[t!]
\centering
\includegraphics[width=0.8\linewidth]{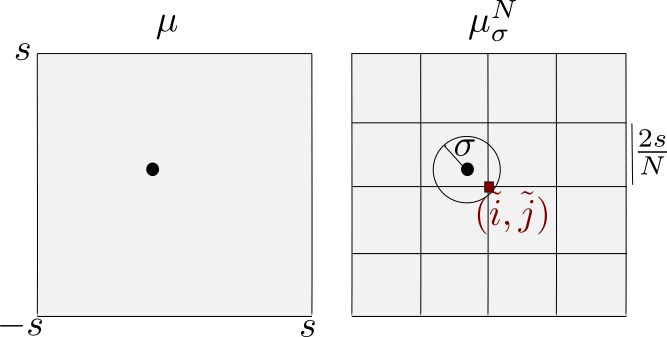}
\caption{\small The differentiable discretization of a courting measure $\mu$ into a point-image $\mu_\sigma^N$ on a $N \times N$ regular grid.
}
\label{fig.disc}
\end{figure}

We are going to approximate our atomic measures on $W_s$ by matrices (images) of given size $N \times N$ (the image resolution), and then use the automatic differentiation software Pytorch \citep{NEURIPS2019_9015} to perform the following operations. 
It allows one to compute the derivative of a modified energy w.r.t. any point $x_i$ in $\mu$. The following paragraph details this discretization:

We first map a given point measure $\mu$ on $W_s$ to a continuous function $\mu_\sigma$ by the convolution 
\begin{equation}\label{e.blur}
     \mu_\sigma(x) := \mu \conv g_\sigma (x) = \sum_{x_i \in \mu}e^{-\frac{|x-x_i|^2}{2\sigma}}   , \quad x \in \W ,
\end{equation}
with a (periodized) Gaussian function $g_\sigma$ of given standard deviation $\sigma$. Then we evaluate $\mu_\sigma$ on the  $N \times N$ regular grid inside $W_s$
and denote the resulting matrix $\mu_{\sigma}^N$, with entries called (values of) {\em pixels}.
The convolution with a Gaussian function makes each entry of $\mu_{\sigma}^N$ smoothly  depend  on the atom positions of $\mu$. We then compute $\bar K(\mu_{\sigma}^N)$ instead of $K(\mu)$,
where $\bar K$ is  this discrete analogy of the descriptor~\eqref{eq:wphK} (cf.~\citet{zm-19}). 
Note that, because the value of a pixel continuously depends on the positions of the atoms, this discretization makes our descriptor only invariant to discrete translations (multiple of the pixel size $2\frac{s}{N}$), for which Theorem~\ref{th.Invariance} applies. The gradient of the energy $|\bar K(\mu_{\sigma}^N) - \bar K(\bp_{\sigma}^N) |^2$ with respect to each atom position of $\mu$ can therefore be computed using automatic differentiation (with the Pytorch software). Indeed, we know that $\bar K$ is differentiable w.r.t. each entry of $\mu_{\sigma}^N$, as a combination of linear and non-linear operators. Moreover, for any $i,j \in \{1,N\}^2$, noting $\tilde i = -s + 2si/N, \: \tilde j = -s + 2sj/N$, \eqref{e.blur} gives us that
\begin{equation}\label{e.discrete}
\mu_{\sigma}^N(i,j) = \mu \conv g_\sigma (\tilde i, \tilde j) = \sum_{x_i \in \mu}e^{-\frac{|(\tilde i, \tilde j)-x_i|^2}{2\sigma}},
\end{equation}
which is differentiable w.r.t. any $x_i$ in $\mu$. This discretization step is illustrated in the Figure \ref{fig.disc}. 

In signal processing, the Gaussian function acts as a low-pass filter. It is needed to cut-off high frequency information of $\mu$ so that $\mu_\sigma$ can be discretized into an image with negligible alisaing effect. This means that $\mu_\sigma$ carries the information on the positions of $\mu$ up to some precision which depends on $\sigma$. The subsequent evaluation of $\mu_\sigma$ on the grid $N\times N$ in $\mu_{\sigma}^N$ implies that
$\sigma$ cannot be taken too small. Indeed, we take $\sigma_{\min}=\frac{s}{N}$ as the lowest value of $\sigma$.

\subsubsection{Wavelet discretization and choice of scales}
\label{sss:waveletdisc}
As stated in section \ref{w_transf}, the family of wavelets used in our descriptor is constructed by dilating the mother wavelet $\psi$ in the range of the scales $0 \leq j < J$. 
Based on the choice of $N$, we set $C = \frac{2s}{N}$. In this way, the spatial support of $\psi$ has a radius $C$ of one pixel of the image. As a consequence, this smallest-scale wavelet $\psi$ can also be discretized (without significant aliasing) in order to compute the discretized descriptor $\bar{K}$.

The choice of the largest scale $J$ can be decided based on the visual structures in the observation. For example, if we want to model structures whose spatial size is close to the size of the window $[0,1/8]^2 \subset \W$, we shall set $ 2^J C = 1 / 8$, i.e. $J=\log_2(N)-3 - \log_2(2s) $.

\subsection{Multiscale optimization}
\label{ss.Multiscale}

Phase harmonic covariance moments of point process images (i.e. point patterns converted into regular pixel grids, as described above) may have large values at high frequencies (large values of $|\lambda|, |\lambda'|$ in~\eqref{eq:wphK}), due to the fact that the point-images are composed of local spikes when $\sigma$ is small. This implies that these high frequency statistics have an important impact on the gradient of $\bar{K}$, which in turn can lead to the gradient descent model being trapped at shallow local minima, where only the high frequencies are well optimized to match the observation. 

This optimization issue can be overcome by matching the descriptors from low frequency to high frequency in a sequential order, through an appropriate
modulation of the parameter $\sigma \in \{\sigma_0, \sigma_1, ..., \sigma_{J-1}\}$ of the Gaussian functions used to discretize $\mu$, introduced in Section \ref{ss.pix}.

Indeed,  since Gaussian functions are low-pass filters, we can interpret the convolution in~\eqref{e.blur} as a blurring, limiting the space localization of Dirac measures.
When such  smoothing of the point pattern is done by a Gaussian function that has a large $\sigma$, the high frequencies of the signal  function are close to 0 and 
the same holds true for the phase harmonics, because wavelets are localized in frequency.
Therefore the wavelet phase harmonics  are dominated by the low frequencies. 
Thus, by smoothing the observed sample and generating the optimal one  with high variance  Gaussian function, we create a new objective leading, in the gradient descent optimization, to a point configuration for which only low frequencies moments (small values of $|\lambda|, |\lambda'|$ in~\eqref{eq:wphK}) are matched with the ones of our observed sample. 
Thus, we propose a \textit{multiscale} gradient descent procedure that consists in choosing first a high value for precision parameter $\sigma$, run the optimization algorithm, and then reduce the value of $\sigma$ to run the optimization again, starting from the result of the previous run (and repeat this operation until $\sigma = \sigma_{J-1}$). We choose $\sigma_j := \frac{s}{N} 2^{J-j-2}$. Note that $\sigma_{J-1}$ is equal to $\sigma_{\min}$.
For numerical efficiency, we perform the gradient descent procedure using the L-BFGS optimization algorithm \citep{l-bfgs}.


\subsection{Final blurring}
\label{ss.Blurring}

We observed that the contrast between the continuous nature of our objects and the discrete approximation described in Section~\ref{ss.pix} creates undesired artificial structures at frequencies higher than the image resolution: when the number of particles in a configuration is large with respect to the number of pixels in the image, or if the configuration exhibits strong clustering behaviour, several pixels may contain more than one particle. In such cases, our algorithm produces samples having an artificial clustering structure {\em inside} each of these pixels (see \citet[Section 5]{brochard2020particle} for an illustration of this phenomenon).

To remove this artificial clustering, we chose to force these high frequencies to be ``as random as possible'', i.e. to have Poisson-like structure. To this end, we introduce a uniform i.i.d. perturbation of the positions of points after the last optimization run. It can be viewed as an additional, this time stochastic,  measure transport, following the deterministic one from the particle gradient descent. 
This final randomization can be viewed as enforcing a-priori information on high frequency structures of the process: Poisson-like structure. 

\section{Numerical experiments}
\label{s.Numerical-Results}

In this section we present numeral experiments involving our generative model.
We begin by presenting in Section~\ref{ss.MAssumptions} our numerical settings,
in particular the distributions of point processes whose samples are used 
as original point patterns. 
 We next evaluate how well our generative model with the phase harmonic covariance descriptor can  generate 
samples similar to those given by the original point processes. 
In Section~\ref{ss.Visual-spcetrum},
 we evaluate these models by comparing samples from the original distributions to samples from our models, visually as well as by estimating their power spectrum, which we define in Appendix~\ref{s.Fourier}. The power spectrum gives information equivalent to the second order correlation function of the process \citep{bremaud2013mathematical}, which captures clustering or repulsive behaviour between atoms of a realization. Such information cannot always be detected visually.
 In order to further quantify how well our model captures visual geometric structures,
 and to gain some insight into the ability of our model to produce diverse samples,
 we shall use the  topological data analysis (TDA), derived from the theory of persistent homology. 
 The comparison will be done in Section~\ref{ss.TDA}. 

\subsection{Numerical settings}

\label{ss.MAssumptions}
We first describe the original point processes that we shall evaluate the particle gradient-descent model, then specify the parameters of the model in the numerical experiments.

\subsubsection{Original point process distributions}
\label{ss.OrignalPP}
For our experiments, we choose point process distributions that show complex geometric structures, for which we can visually recognize geometric structures.
We begin by presenting results for 
{\em Cox  (double-stochastic Poisson) processes} with Poisson points living on one dimensional structures generated by two famous stochastic  geometric models, namely edges of the Voronoi tessellation, see e.g. \citet{skare2007bayesian}, and the Boolean model with circular grains of fixed radius, considered in \citet[Example 10.6]{chiu2013stochastic}. Both underlying geometric models are generated by a Poisson parent process within the observation window $W_s$, and we construct these models in a periodic way to avoid border effects.
We call the  respective Cox processes {\em Voronoi} and {\em Circle} processes. 
Note that, for these two processes, Poisson points live on different geometric shapes: polygons for the Voronoi and possibly overlapping circles for other  one. Additionally, we consider two different radii of circles.

Then, we take interest in distributions having {\em turbulent intensity} (derived from the simulations of a decaying isotropic turbulent vorticity field driven by 2d Navier-Stokes equations, see e.g. \citet{schneider2006coherent}). Such fields  exhibit complex mulsticale structures, and are representations of physical phenomena, known to be difficult to model faithfully. Furthermore, the distributions we consider have much greater intensities that the previous Cox models.
From the turbulent intensity, we sample three different processes, exhibiting distinct microscopic structures (repulsive, independent or clustering): a  Matern {\em cluster} process, a Poisson point process and a Matern II {\em hard-core} process, see \citet[Example 5.5 and Section 5.4, respectively]{chiu2013stochastic}. We study the ability of our model to reproduce simultaneously the macroscopic (i.e. the turbulent intensity) and microscopic structures (i.e. at small scales) of the process.

The number of points in the Cox Voronoi, Small circles, Big circles, and the Turbulent Hardcore, Poisson, ad Cluster processes are around, respectively, 1~900, 2~500, 2~000, 1~700, 3~800 and 13~000.
Note that, for comparison, point patterns considered in \citet{Tscheschel2006} have around 400 points.

\subsubsection{Choice of model parameters}
\label{ss.it}

\paragraph{Image resolution}
As discussed in Section \ref{ss.pix}, point configurations are convoluted with Gaussian densities 
and evaluated on $N \times N$ grid (as images) in order to efficiently compute our descriptors, and move the particles with gradient descent.
For simplicity, we fix $s=1/2$ for all the examples that we shall consider.
The ultimate Gaussian variance (precision) of this mapping is thus $\sigma_{\min}=\frac{1}{2N}$.
The larger $N$ is, the more information we are able to keep (in high frequencies), but the larger the computation time. 
We chose for our experiments a resolution of $N = 128$. We show one example where a higher resolution, $N=256$,  is used  to capture most of the high frequency information.

\paragraph{Number of iterations}
The number of iterations of the L-BFGS optimization is chosen to be $100$ for each scale $\sigma_j$ 
(a total of 400 iterations for $N=128$, and 500 iterations for $N=256$). 

\paragraph{Other parameters and computation time}
Empirical evidence in Section~\ref{ss.Visual-spcetrum} shows that the multi-scale optimization procedure in Section~\ref{ss.Multiscale}
allows one to reconstruct (modulo translation) the observed sample when using $\bar K$ defined with $J=\log_2(N)-2$, which is not the case when simultaneously optimizing all frequencies. In order to preserve the ability to reproduce geometric structures at all scales, we shall also apply this multiscale optimization method to our model defined with $J = \log_2(N)-3$.
The number of angles in the steerable wavelets is $L=8$.

An overview of the main parameters of our model is given in Appendix \ref{s.params}. 
The average computation time on 4 GPU (Nvidia Tesla P100) 
for a sample  for a turbulent process having roughly 13~000 points with resolution $N=256$ is between 5 and 10 minutes while the same task at the resolution $N=128$ takes  between 1 and 2 minutes.

\subsection{Visual evaluation and spectrum comparison}
\label{ss.Visual-spcetrum}

\begin{figure*}[h!]
\centering
\begin{tabular}{r@{\hskip0.5em}c@{\hskip0.5em}c@{\hskip0.5em}c@{\hskip0.5em}}
&Voronoi &Small circles&Big circles\\
 \rotatebox[origin=l]{90}{\hspace{3em}Original}&
 \includegraphics[width=0.25\linewidth]{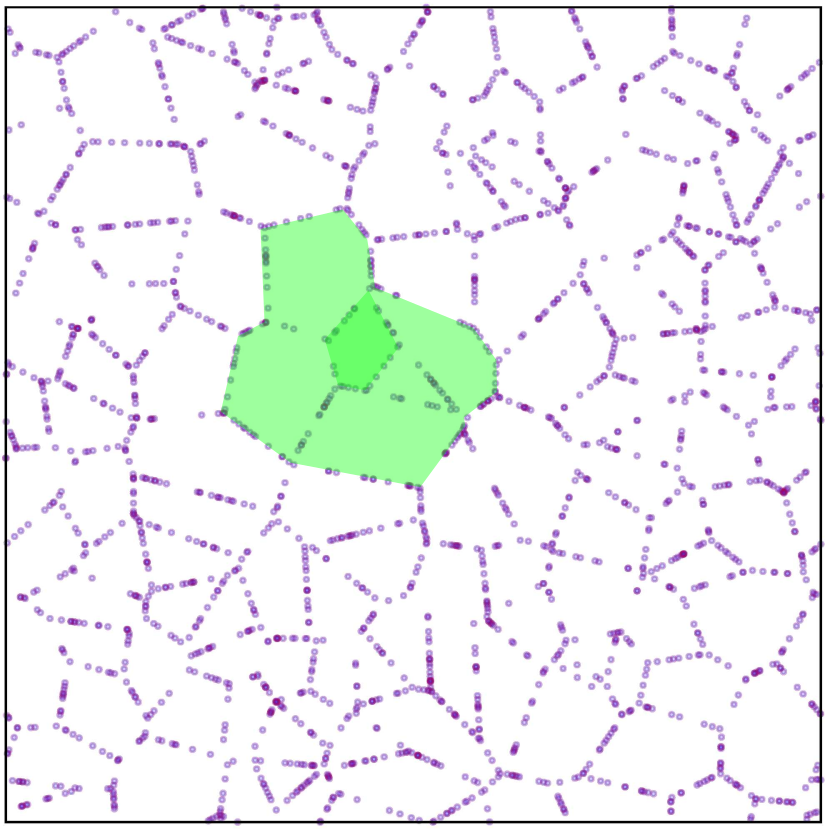}&
\includegraphics[width=0.25\linewidth]{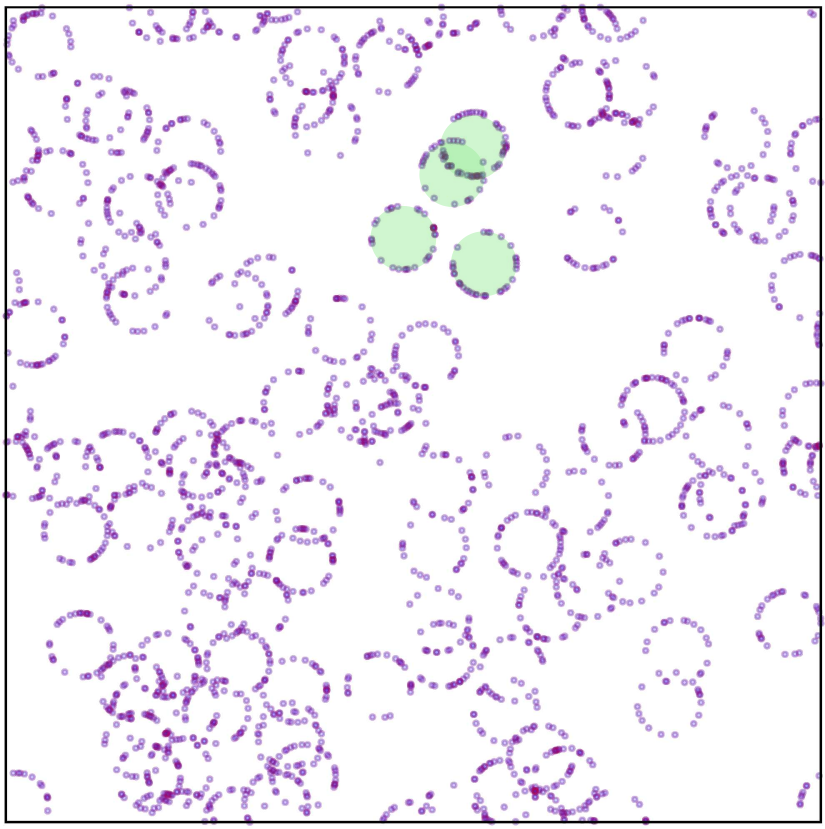}&
\includegraphics[width=0.25\linewidth]{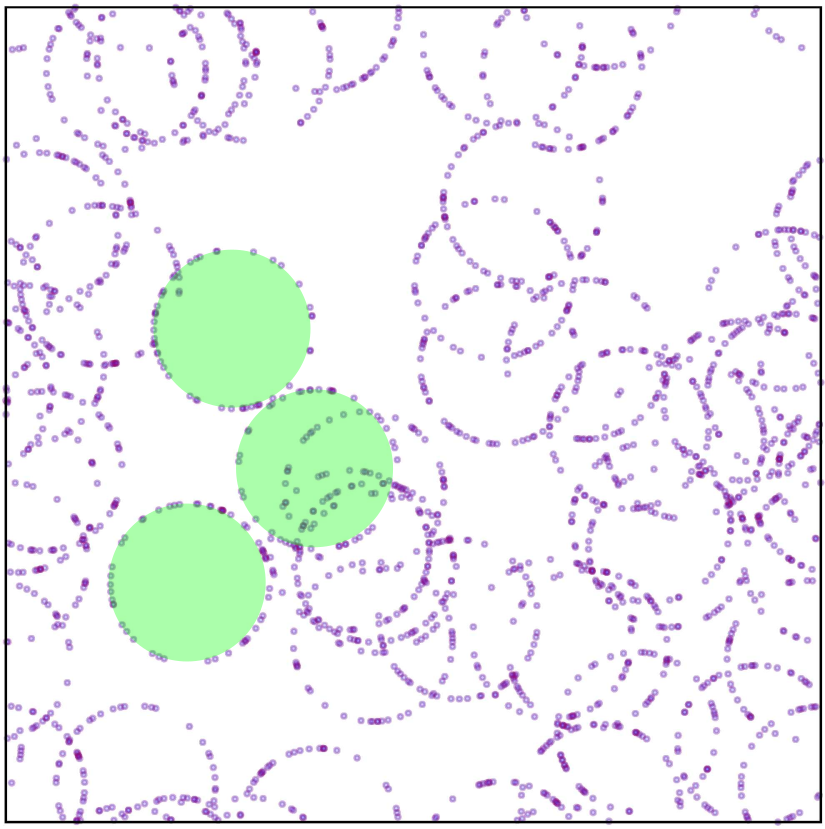}\\
 \rotatebox[origin=l]{90}{\hspace{1em}Reconstruction}&
\includegraphics[width=0.25\linewidth]{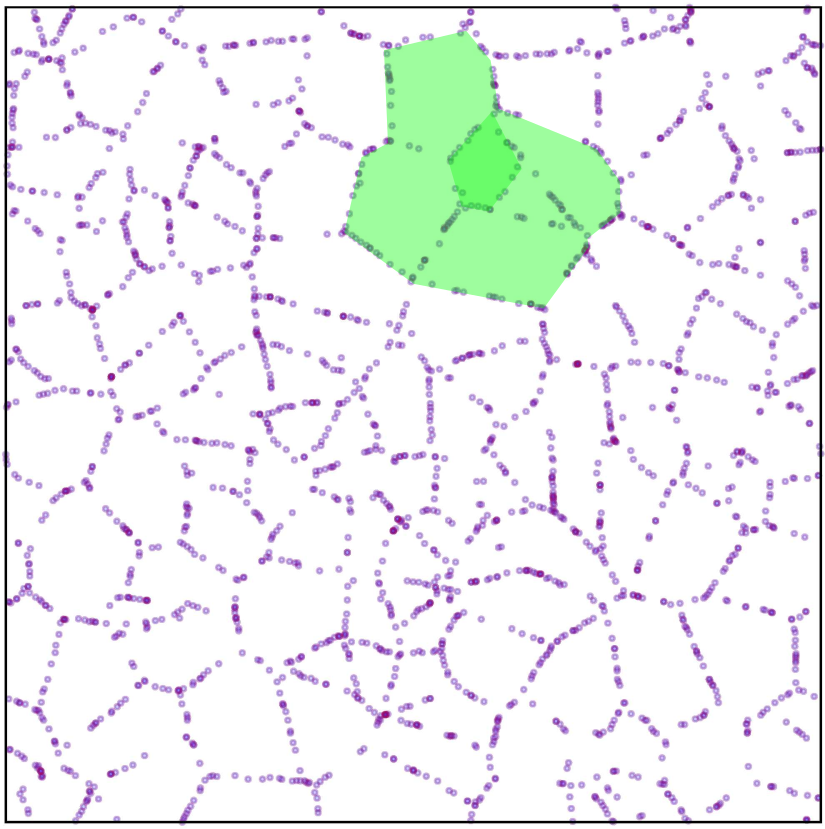}&
\includegraphics[width=0.25\linewidth]{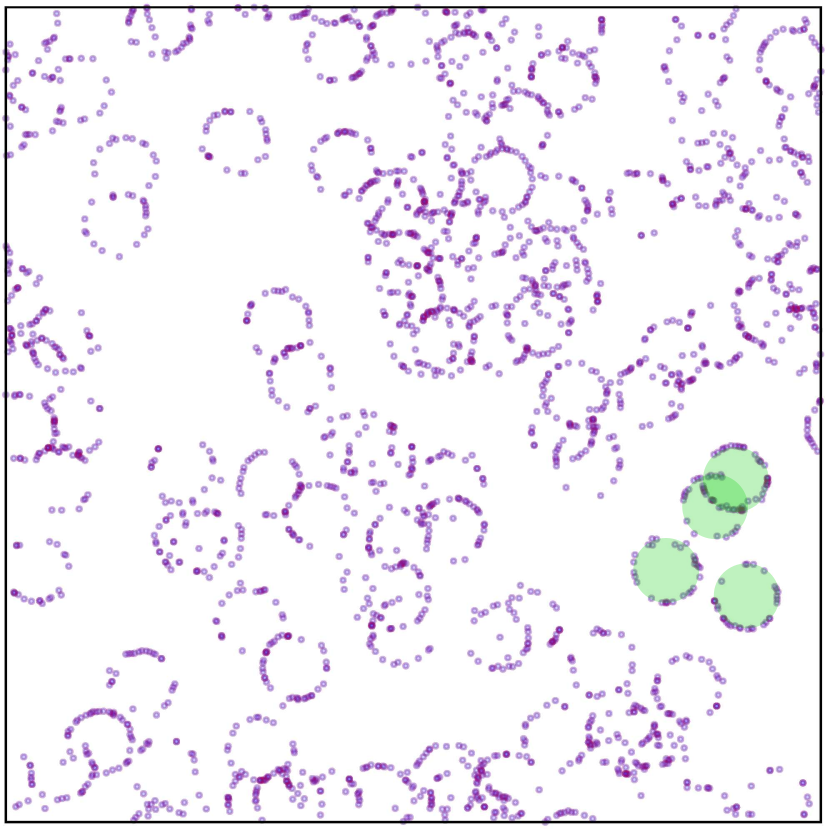}&
\includegraphics[width=0.25\linewidth]{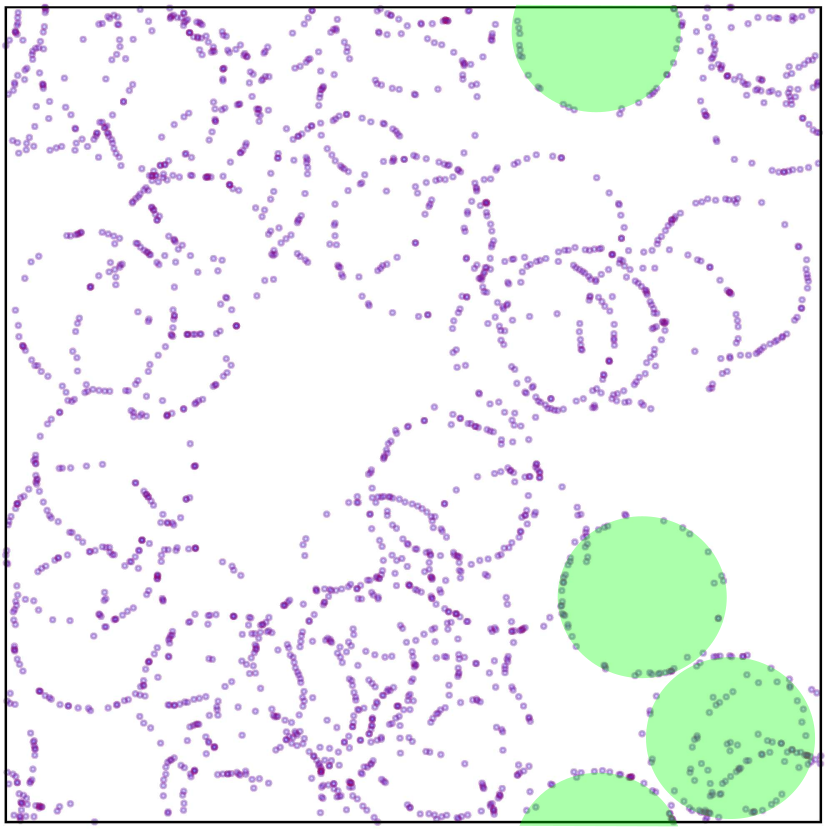}\\
\rotatebox[origin=l]{90}{\hspace{3em}Synthesis}&
\includegraphics[width=0.25\linewidth]{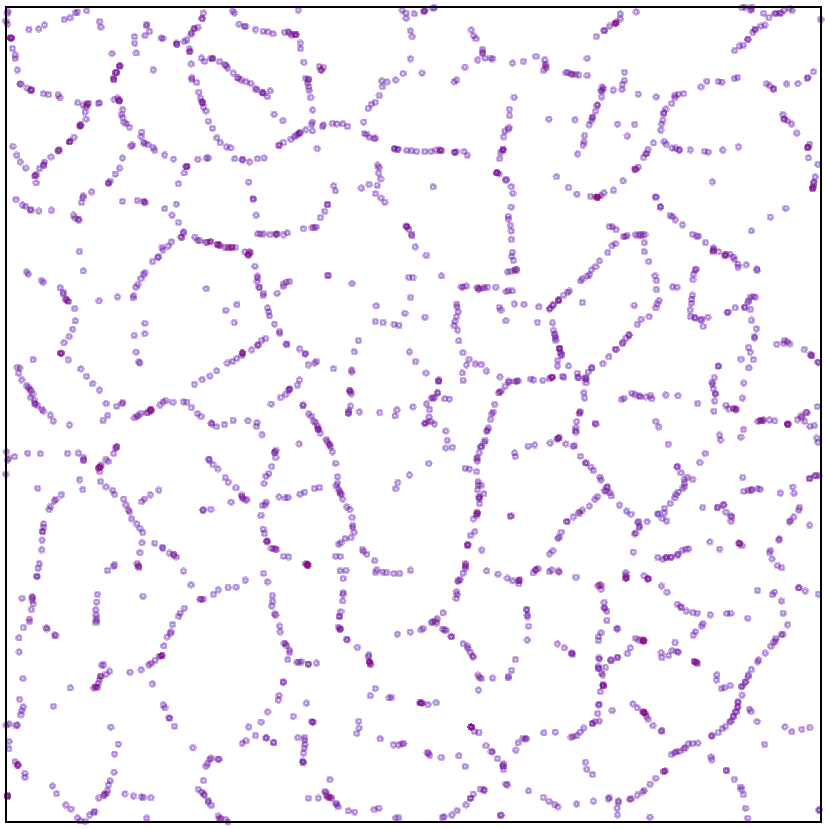}&
\includegraphics[width=0.25\linewidth]{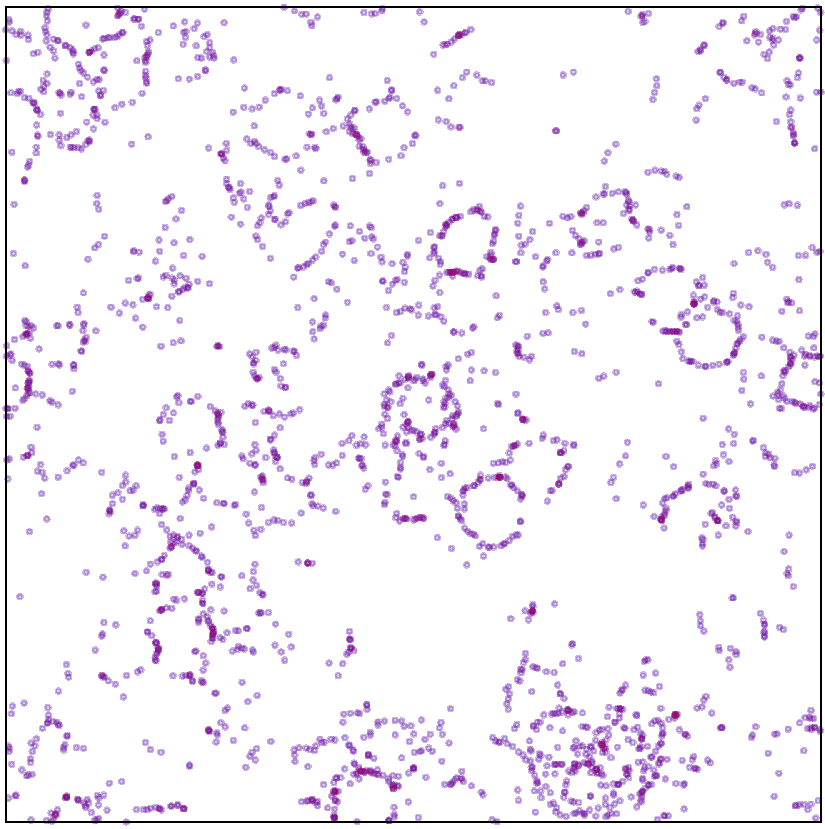}&
\includegraphics[width=0.25\linewidth]{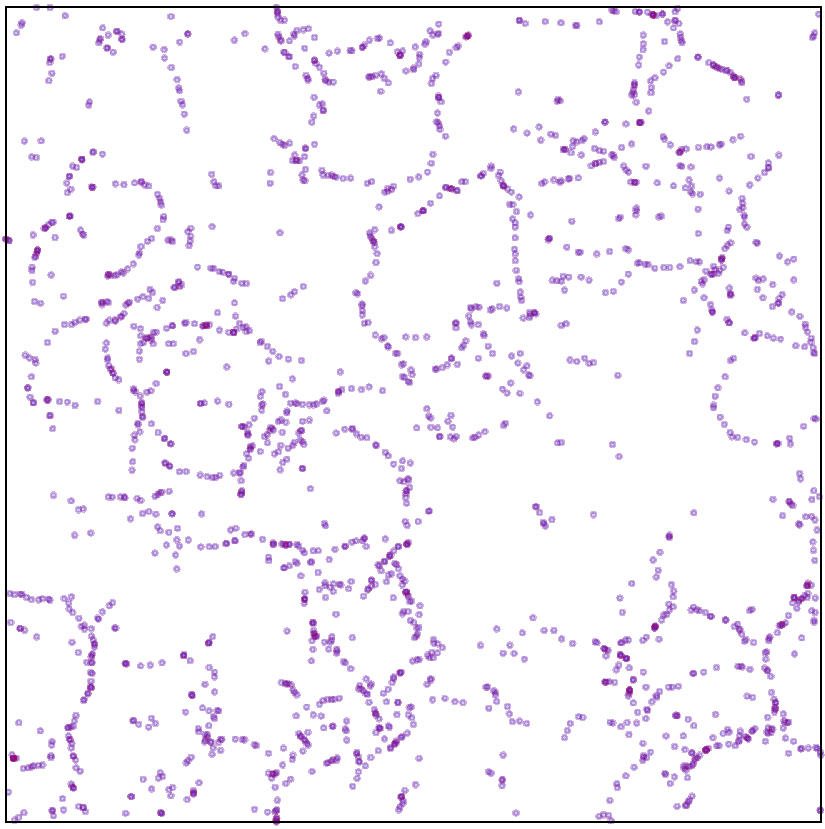}\\[2ex]
\rotatebox[origin=l]{90}{\hspace{1em}Power spectrum}&
\includegraphics[width=0.28\linewidth]{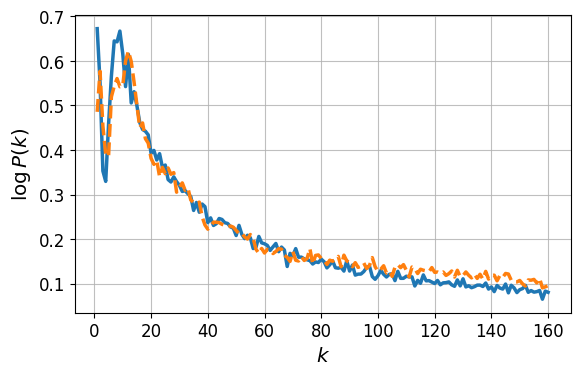}&
\includegraphics[width=0.28\linewidth]{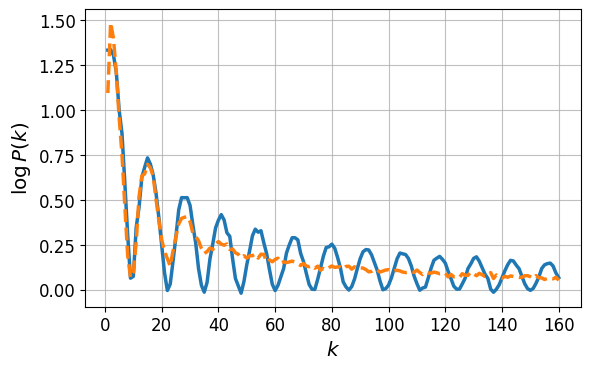}&
\includegraphics[width=0.28\linewidth]{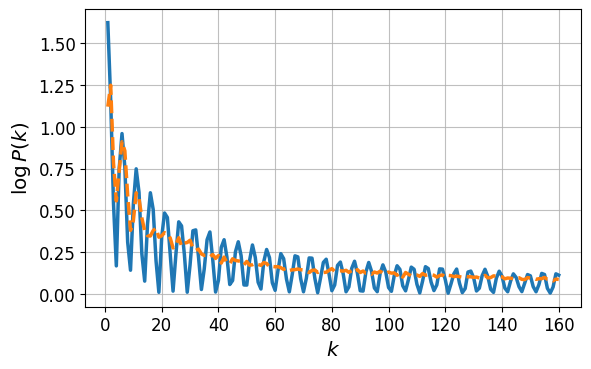}\\
\end{tabular}
\caption{\small Three Cox models; original sample, reconstruction, and synthesis. For the power spectrum plots, the full lines (in blue) correspond to the original distributions, and the dashed lines (in orange) correspond to the models. The y-axis is presented in log scale.
\label{fig.geometric}}
\end{figure*}

\begin{figure*}[t]
\centering
\vspace{-2ex}
\begin{tabular}{r@{\hskip0.1em}c@{\hskip0.5em}c@{\hskip0.5em}c}
&Hard-core&Poisson&Cluster\\
\rotatebox[origin=l]{90}{\hspace{3em}Original}&
\includegraphics[width=0.25\linewidth]{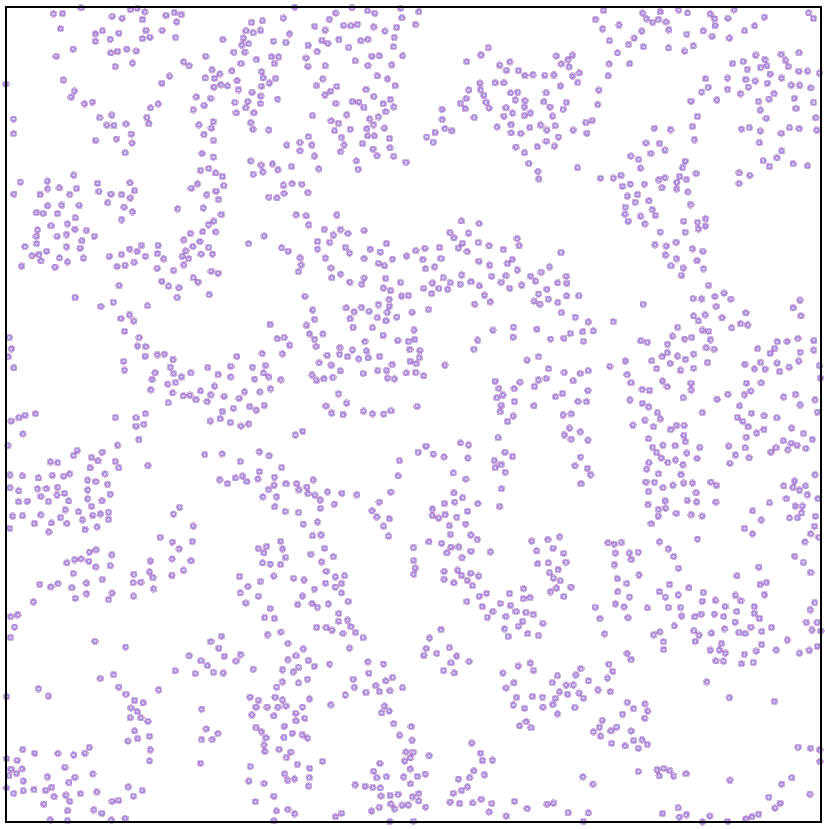}&
\includegraphics[width=0.25\linewidth]{Figures/Original_data/turb_zoom_sparse}&
\includegraphics[width=0.25\linewidth]{Figures/Original_data/turb_zoom_cluster0-s}\\
\rotatebox[origin=l]{90}{\hspace{3em}Synthesis}&
\includegraphics[width=0.25\linewidth]{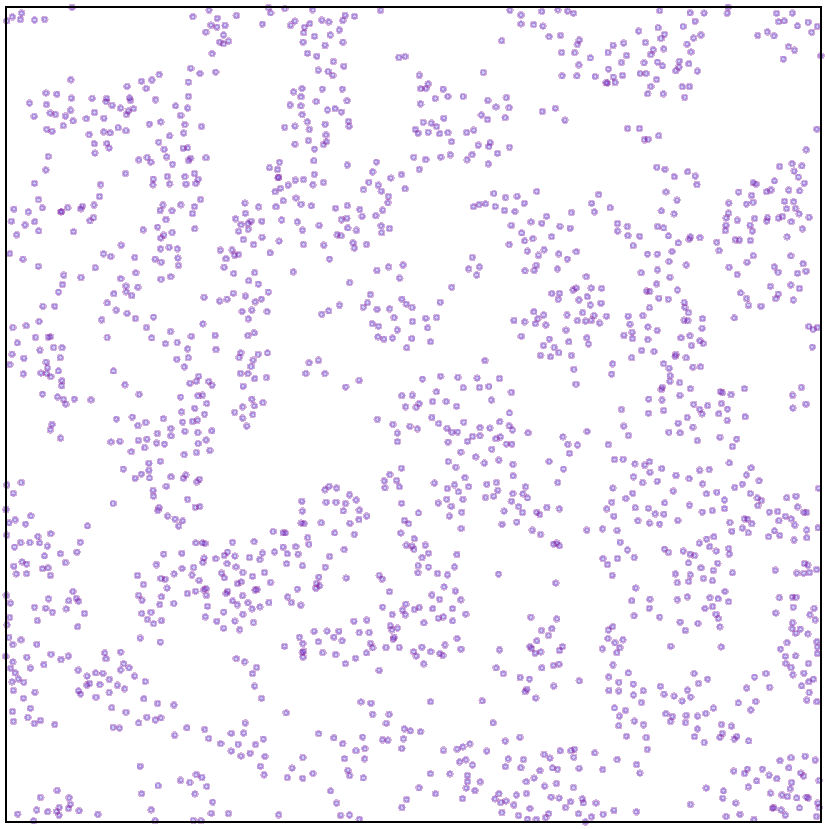}&
\includegraphics[width=0.25\linewidth]{Figures/Syntheses0307/turb_zoom_sparse_phsyn1}&
\includegraphics[width=0.25\linewidth]{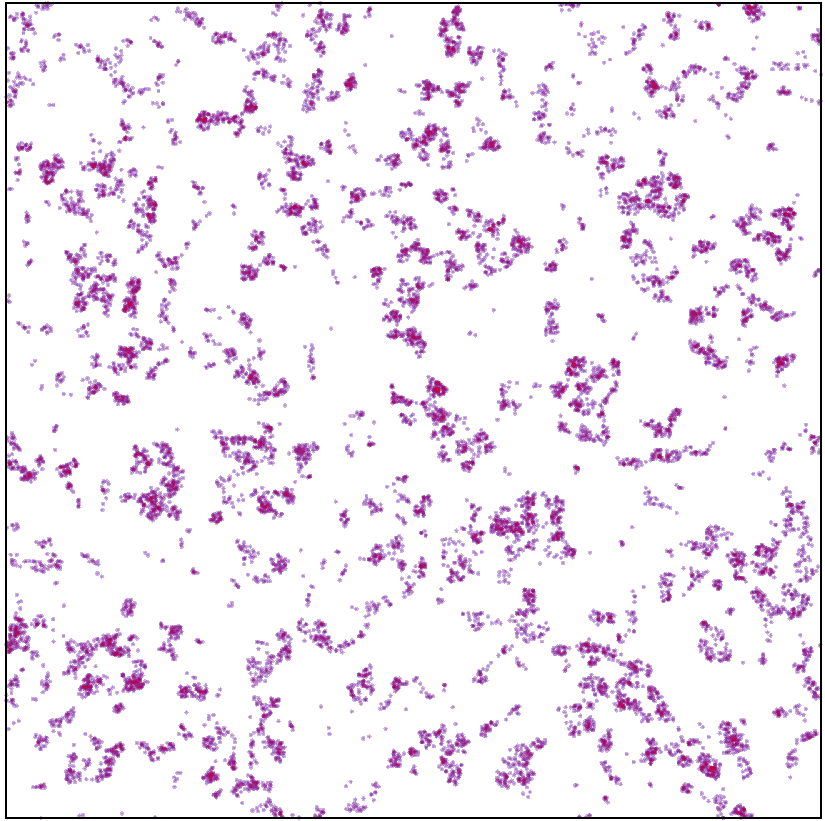}\\[1ex]
\rotatebox[origin=l]{90}{\hspace{1em}Power spectrum}&
\includegraphics[width=0.28\linewidth]{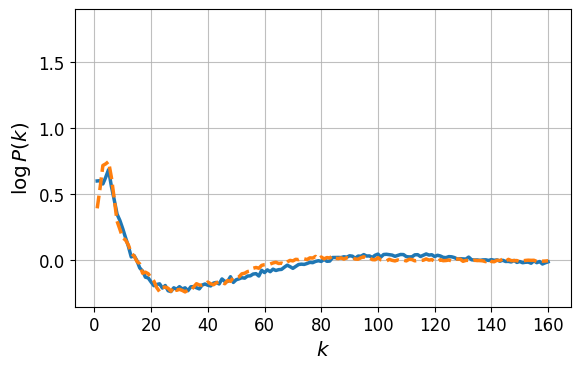}&
\includegraphics[width=0.28\linewidth]{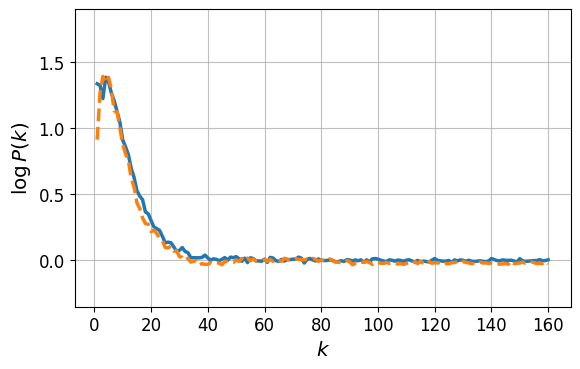}&
\includegraphics[width=0.28\linewidth]{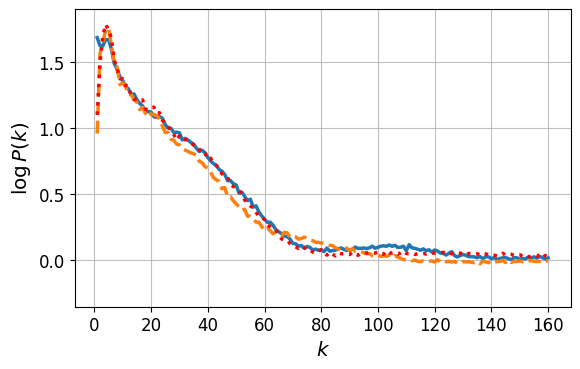}
\end{tabular}
\caption{\small Synthesis of turbulence processes with various microscopic structures (Hardcore, Poisson, and Cluster), and their power spectrum plots. Full lines (in blue) correspond to the original processes, dashed lines (in orange) correspond to the model. For the Cluster distribution, the dotted line (in red) corresponds to our model defined with the resolution $N=256$.  \label{f.PoissonIII}}
\end{figure*}

We evaluate the ability of our model to capture and reproduce geometric structures exhibited by realizations of the point processes described in Section \ref{ss.MAssumptions}. 
A natural first method to assess the sufficiency of a generative model (property (P2)) is visual evaluation, which is widely used in image analysis but subjective. 
We then compare the power spectra (cf. Appendix \ref{s.Fourier} for the definition) of our models and the original distributions.
To estimate the power spectra, we generate (for each original distribution) 10 i.i.d. samples from the same model (i.e. from the same observation sample $\bp$ , but with different initial configurations $\bp_0$). We average the power spectra of the 10 syntheses, and compare it to the average of 10 i.i.d. samples from the original distribution. All these samples will also serve in Section \ref{ss.TDA} to compare their geometric similarities.

Figure~\ref{fig.geometric} shows a study of our three Cox distributions. The first line presents samples from the original distributions.
The second line presents samples from the model using our descriptor with $J = \log(N)-2$.
In this setup, the concentration property is not satisfied (see property (P1) in Section \ref{sss.maxent}), and the result is the memorization of the observation sample $\bp$.
Indeed, this line shows quite faithful reconstructions of the original samples subjected to a periodic translation, up to some precision error due to a finite image resolution $N$. This is however not a good model because it essentially only contains the observation $\bp$. It suggests that we need to improve the concentration property (P1) of the descriptors in order to enlarge the ensemble $\Omega_\epsilon$. Note that, in the work of \citet{Tscheschel2006}, the authors use the term 'reconstruction' to refer to random sampling method, which we call in this paper 'synthesis'.  

In order to improve (P1), we shall reduce the parameter $J$ in the wavelet transform. The third line of Figure~\ref{fig.geometric} shows realizations sampled using 
$J=\log(N)-3$ for different original distributions. 
Our analysis in Section \ref{sss:waveletdisc}
suggests that this range of $J$ can model structures 
whose spatial size is at most 1/8 of the window $W_s$. 
Observe that most polygons and circles are well reproduced in the synthesis of Voronoi and Small circles. The Big Circles are harder to model since the size of each circle is slightly larger than  1/8 of $W_s$.

In the last line of Figure~\ref{fig.geometric}, we present the power spectra for $k \in \mathbb N \cap [1,128[$ (cf. Appendix~\ref{s.Fourier})  from the original distributions and as well as from our model. Larger errors can be observed at $k$ near zero (say k=1,2 and 3). This is because only 
the average spectral information is captured (and matched) using the low-pass filter $\psi_0$ in the wavelet transform, 
which is included in the descriptor $K(\mu)$ (c.f. \eqref{eq:wphK}). Moreover, 
the variance of the empirical information at small $k$
can create extra error since it can be far away from its expectation. 
Similarly, because the wavelet convolutions average the spectral information over 
different frequency bands when using a reduced number of $\tau'$ in $K(\mu)$, 
our descriptor does not capture fast oscillations in the power spectra in the range of $k \leq N/2 =64$  (see \citet{zm-19} for more details about how to capture these oscillations).
This is observed in the cases of Small and Big Circles Cox processes. See \citet[Appendix C]{brochard2020particle} for a theoretical formula of the power spectrum in the case of Small and Big circles. When $k > N/2=64$, the 
descriptor $\bar K$ (cf. \ref{ss.pix}) does not contain accurate spectral information due to a finite image resolution $N$. In this regime, we observe a smooth decay of the (log) power spectrum towards 0. We observed that if we apply 
the final blurring (cf. Section~\ref{ss.Blurring}), then the error of the model spectrum becomes larger. Therefore, for these three processes, no final blurring has been applied.

All models discussed up to now are Cox processes, with Poisson (hence independent) points sitting on some random macroscopic structures.
Figure~\ref{f.PoissonIII} presents our analysis of three turbulent point processes having different microscopic structures: a hardcore, a non-correlated (Poisson) and a clustering one. We see that our generated samples capture to some extent this microscopic structure. For the clustering model, the presented synthesis is done with $N=256$. 
We see that our model (using $J=\log(N)-3$) can generate samples with similar macroscopic and microscopic structure. The power spectrum at small $k$ has larger errors, as we have observed in the Cox models. However, since the power spectra are mostly smooth in these Turbulent processes, we observe a relatively small spectrum error over a wide range of $k$. This is also due to the use of the final blurring which helps to remove some artificial spectrum errors for $k > N/2$. 
For the clustering model, we also compare the power spectrum of two models with different resolutions: $N=128$ and $N=256$. We see that setting a higher resolution reduces significantly the error, allowing to match the spectrum up to $k \simeq 80$.
we still observe some small error when $ k \geq 80$, probably because of the final blurring (cf. Section~\ref{ss.Blurring}), which may also impact the high frequencies that we optimize. Overall, both the visual and the spectral analysis suggest that our model can generate well various Turbulent points processes.

\subsection{Persistent homology and topology analysis}
\label{ss.TDA}

As previously mentioned, power spectrum evaluation corresponds to the comparison of second order moments, which only partially capture geometric structures. Visual evaluation can be more discriminate, but is subjective. 
To evaluate more precisely the ability of our model to capture the geometric structures of the given distributions, we shall use a representation of objects derived from persistent homology theory,
which is a powerful algebraic tool for studying the topological structure of shapes, functions, or in our case point clouds. We shall perform this evaluation by comparing the {\em persistence diagrams} of the generated samples to those of the original ones. Furthermore, this representation allows us to evaluate in a simple way the ability of our model to produce diverse samples.

We begin by a brief, intuitive presentation of persistence diagrams, and the whole comparison method that will be simply referred to as topology data analysis (TDA).
For more details we refer the reader to \citet[Section~11.5]{boissonnat2018geometric}. 
We then present the TDA of our point process distributions and models.
TDA can be seen as a complementary tool with respect to the spectrum analysis,
being more consistent with visual perception (see \citet[Appendix C]{brochard2020particle} for more details about this link).

\paragraph{Persistence diagram}

Persistent homology theory describes a way to encode the topological structure of a point cloud through a representation called  {\em persistent diagram} (PD). It is constructed,   for a given  point configuration $\phi \in \mathbb M^{s}$, from the family  $(G_r)_{r\geq0}$ of Gilbert graphs, where the vertices are the positions of atoms of $\phi$, and the edges are pairs of points closer to each other than $r$.\footnote{In our case we use the periodic metric.}  Then, we fill-in the triangles (triplets of points joined by edges) of the graph. Points, edges and filled-in triangles
constitute the so-called 2-skeleton of the Vietoris-Rips (VR) complex.
For any $r\geq0$, we study two characteristics of the skeleton: its {\em connected components}, and its {\em holes} (this latter notion is well formalized in the algebraic topology, in our case  they correspond to the natural idea of a hole).
Each connected component ``is born'' at  time (radius) $r=0$ and it ``dies'' at some time $r>0$ when it is merged 
with another connected component. Similarly, each hole has a  birth time ($r>0$) corresponding to the minimal radius at which it appears,
and a (larger) death time corresponding to the minimal radius for which the hole is completely filled-in by the triangles.
The persistence diagram of $\phi$
is the collection of  pairs of birth and death times of the connected components and holes. 
It is hence a point process in the positive orthant of the plane, offering a multiscale  
(as our wavelet-base descriptor)  description of the topology of $\phi$. As our descriptor, it is also stable to small deformations of~$\phi$. It is hence interesting to use this alternative tool to evaluate our generative model.

\paragraph{Topological data analysis}

Our approach in this matter is inspired by \citet{tda}, and we refer the reader to this paper for a more detailed description. We use the 'holes' birth-death process, as it appears more relevant to capture information in the Cox distributions, such as the polygons and the circles.

In order to compare the distributions of our models to the original distributions,
we compute the PDs of our samples from each distribution (cf. Section \ref{ss.Visual-spcetrum} for a description of these samples). Recall, these  PDs can be viewed again as point clouds in two dimensions. Therefore, a distance between two PDs can be computed, and we use in this regard a periodic version  of the Wasserstein distance
between two point clouds on the plane (we found that the bottleneck distance, also suggested in \citet{tda}, is not sufficiently discriminating for our point patterns). 
We obtain in this way a distance matrix between different PDs (reflecting topological similarities or differences of the point processees realizations for which PDs were calculated).  
We then apply a standard dimension reduction algorithm (namely Multi Dimensional Scaling)
to this distance matrix, to represent every PD (and hence the corresponding sample) as one point on the plane, and we visualize the representation of all samples.

\begin{figure*}[h!]
\centering
\begin{tabular}{r@{\hskip0.5em}c@{\hskip0.5em}c@{\hskip0.5em}c}
&Cox Voronoi &Cox Big circles &
Turbulence hardcore\\
&\includegraphics[width=0.25\linewidth]{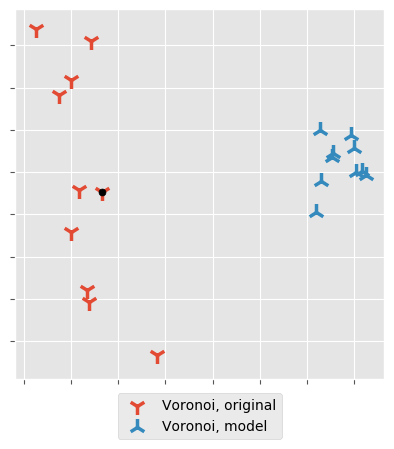}
&\includegraphics[width=0.25\linewidth]{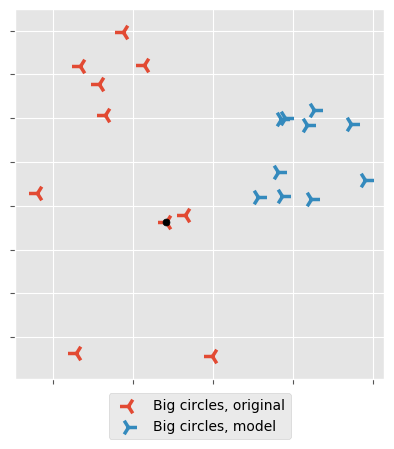}
&\includegraphics[width=0.25\linewidth]{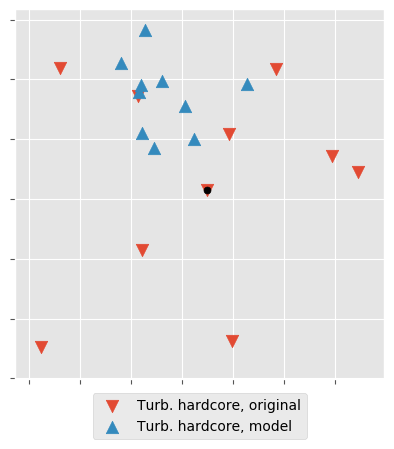}\\[1ex]
\end{tabular}
\begin{tabular}{c@{\hskip2em}c@{\hskip0.5em}}
Cox --- three distributions &Turbulence --- three distributions\\
\includegraphics[width=0.405\linewidth]{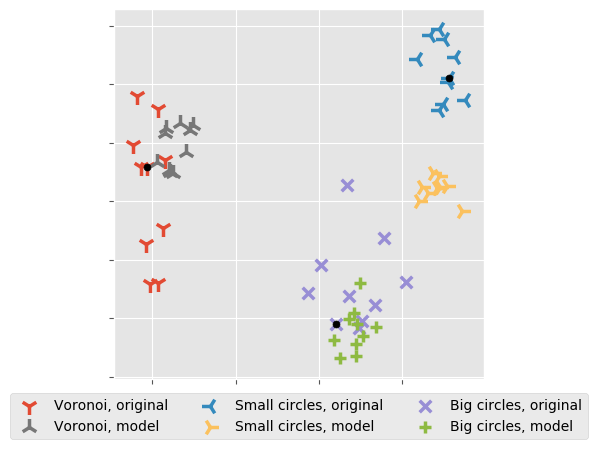}
&\includegraphics[width=0.37\linewidth]{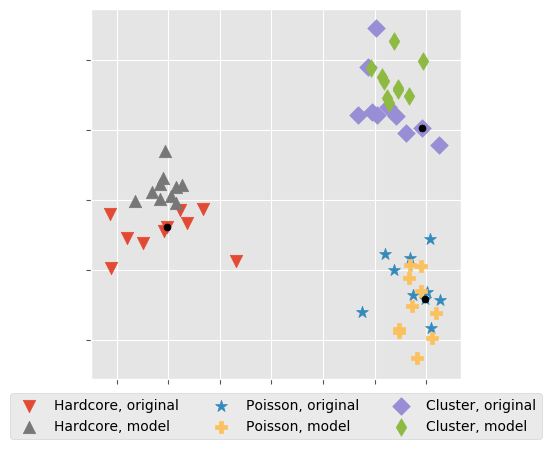}
\end{tabular}
\caption{TDA of the distributions presented in this paper, and their respective models.
\label{fig.TDA}
}
\end{figure*}

\paragraph{TDA  of our experiments}

In the plots on the  first line of Figure~\ref{fig.TDA}, we study separately the Cox Voronoi and Big circles processes, and the turbulent hardcore process. In each plot, we observe 20 {\em dots} (having different shapes), 
each representing one configuration of points in $W_s$ (the term "dot" is used to avoid confusion with points in $W_s$).
For each model there are 10 dots representing i.i.d. realizations of the original distribution and 10 representing realizations from the generative model. For each plot, the sample additionally marked with a black dot represents the observation used in our model to produce the 10 syntheses.

We see in the first two Cox examples a clear separation between the original process and the model, implying a lack of sufficiency (P2) in our model.
This is probably because, in order to satisfy (P1) and produce diverse samples, we have chosen to reduce $J$, and therefore lose some information about large scale structures of the process. On the other hand, we observe that this separation is smaller for the Turbulence hard-core case, where there is a better balance between (P1) and (P2). The error in the Cox models is probably due to the difficulty to reproduce highly constrained structures (perfect circles or convex polygons). These observations agree with our visual evaluation of the syntheses: the Voronoi and (more particularly so) the Big circles models are easily discriminated from their original distributions (their highly constrained structures are not perfectly reproduced in the syntheses), but this discrimination is harder for the Turbulent hardcore case. Moreover, these figures indicate, by the spread of the dots representing the syntheses, that our model reproduces, to some extent, the diversity in the samples of the original distributions  (suggesting a certain entropy in our model). To further reduce the distances between the model samples and the original samples of a process, while maintaining a similar diversity (and hence a similar entropy between the model distribution and the original distribution) remains an interesting problem for future works.

The two plots in the second line present the TDA of the three Cox distributions together, and the three turbulent distributions together.  We observe that, for the Cox distributions, the Small circles model is about as close to the original Small circles distribution as it is to the original Big circles distribution. 
Nevertheless, the original distribution of the Big circles and the Small circles are well separated, suggesting that there is a topological distinction that is not well respected in the small circles. However, this kind of error is hard to perceive visually. 
On the other hand, the Voronoi case is well separated from the other two, suggesting that the model is better than the ones of the circles distributions. For the turbulence case, we observe that the three distributions (both original and model) are well separated, and each respective model is closer to its original distribution than to the other distributions. This agrees with our earlier visual and spectral analysis, suggesting that our model is able to capture complex geometric structures formed by a large amount of points. 

It remains an open question to quantify the influence of the range parameter on the distances between patterns. We chose to include all radii ($0 \leq r \leq \frac{1}{2}$), as we want to include information pertaining to individual point patterns, in order to measure the diversity in the distribution.
In order to get an idea of the influence of this range parameter, it is possible to look at the Euler-Poincar\'e characteristic (see e.g. \citet[Chapter 4]{illian}), which is closely linked to TDA (See Section \ref{s.comparison} for further discussions). Other related methods to compute distances between point patterns, such as in \citet{muller2020metrics}, could constitute an interesting line of research for other evaluation methods.

\textbf{Remark:} We used the R packages {\tt TDAstats} \citep{wadhwa2018tdastats} to calculate the PDs of our point patterns and  {\tt TDA} \citep{fasy} to calculate their Wasserstein distances. Due to memory constraints, for second line, the analysis was done using a random thinning to reduce the number of points of each sample to 2~000, which could artificially impact the results. The experiments were repeated several times and the variability in the random thinning did not impact our conclusions.

\section{Comparison between our method and \citet{Tscheschel2006}}\label{s.comparison}

In this section, in order to illustrate the advantages of the method presented in this paper, we present a brief comparison between the method presented in \citet{Tscheschel2006} and ours.

\subsection{Differences between the two methods}

Both methods are based on the following idea: to produce similar but different point patterns to a given observation, one first defines what should be 'similar' between the observation and the synthesis, by choosing a set of statistical constraints, computed on the observation. Then, starting from an initial random configuration of points, one iteratively modifies this configuration in order to match the set of prescribed statistics (by minimizing an energy $E$, related to the square difference between the statistics of the original and the synthesised  point patterns). If the set of statistics does not describe the observation itself, but rather its underlying distribution, then the output of the optimization procedure should be a new point pattern, similar but different to the observation.

However, the two methods differ on two major points:

\begin{itemize}
    \item First, the optimization method to match the set of statistics. In \citet{Tscheschel2006}, the optimization steps can be described as follows: given the point configuration being synthesized $\displaystyle{\phi_k = \sum_{i=1}^N \delta_{x_{i,k}}}$ at some step $k$, a point in the configuration is chosen uniformly at random, say $x_{j,k}$, for $j \in \{1, \cdots, N \}$. A candidate for a new point $y \in W$ is chosen uniformly at random in the observation window. Then, if the energy of $\tilde \phi_k := \phi_k - \delta_{x_{j,k}} + \delta_y$ is lower than the energy of $\phi_k$, we define $\phi_{k+1} = \tilde \phi_k$. Otherwise, $\phi_{k+1} = \phi_k$. We call this optimization {\em random search} (RS).
    
    In our method, $\displaystyle{\phi_{k+1} = \sum_{i=1}^N \delta_{x_{i,k} - \nabla_{x_{i,k}}E(\phi_k)}}$, where $\nabla_{x_{i,k}}E(\phi_k)$ is the gradient of the energy with respect to the point $x_{i,k}$ of $\phi_k$, see \eqref{e.particle-gradient}. This optimization method will be noted (GD).
    \item Second, the set of statistical constraints used to describe the geometry of the point patterns. In \citet{Tscheschel2006}, the authors use the $k$ nearest neighbour distance distribution functions (d.f.) $D_k(r)$, for $k \in \{ 1, \cdots, k_{max} \}$, $k_{max} \geq 1$, and evaluated at a sequence of radii $r \in \{r_0, \cdots, r_{max}\}$, $r_{max}>0$, \citep[p. 267]{stoyan1994fractals}. We call this statistical descriptor {\em nearest neighbour distances} (NND). 
    
    Our statistics are based on the covariance between phase harmonics of the wavelet phase harmonics coefficients (WPH) of the point patterns (see \ref{s.Wavelets}).
\end{itemize}

\subsection{Preliminary  discussion}

Before presenting a numerical comparison between the two methods, we briefly explain what the limitations of the method in \citet{Tscheschel2006} are, and why our method could overcome such limitations.

The optimization method in \citet{Tscheschel2006} is based on random search. This implies that for some configuration $\phi_k$ at some step $k$, there may be a lot of failing new candidates to replace some point in $\phi_k$ before finding one that reduces the energy of $\phi_k$. This means that there may be a lot of energy evaluations before updating the current configuration. Furthermore, each iteration, requiring one energy evaluation, only moves one point in the configuration. Conversely, at each iteration,  our algorithm computes the energy as well as the gradient of the energy, and all the points are moves according to the gradient. This implies that, for some energy level $e>0$, the gradient descent method may reach this level in less iterations.

Moreover, the statistical constraints used in \citet{Tscheschel2006} are based on 3 parameters: the number of neighbours for the points in the configuration, the maximal radius at which to evaluate whether or not there is a neighbour, and the number of radii between 0 and this maximal radius.
While the latter relates to the precision of the d.f.'s, $k_{max}$ and $r_{max}$ have to be chosen carefully, so as to describe the geometry formed by the points, up to some scale.
The maximal radius $r_{max}$ can be seen as the maximal scale up to which the constraints describe the geometric structure. This can be fixed depending on the observation, but should not be too large, in order to satisfy the ergodic averaging property. However, for a fixed sequence of radii, the parameter $k_{max}$ can change significantly depending on the observation. Even if two configuration exhibit structures up to similar scales, the number of nearest neighbours inside some ball may differ depending on the intensity of the process. 
For instance, consider the Cox Circles distribution, where points are located on circles of fixed radius $r_0$, with the center of those circles forming a Poisson point process. If the observed pattern has around 10 points per circle, one would probably need to fix $r_{max}=2r_0$, and $k_{max}=10$. However, if the circles contain an average of 100 points, then one would have to increase $k_{max}$ up to 100, even if the circles have the same size as before. This would increase significantly the number statistics to compute at every step. Conversely, our descriptors only depend on a number of scales at which we compute the wavelet coefficients, which does not depend on the intensity of the process. The only parameters to fix are the maximal scale $J$, and the minimal scale, set by the resolution $N$, which relates to the precision set by the number of radii in the $k$ nearest neighbours case. In the above Cox Circles example, if the points have Poisson distribution on the circles, then the size of the descriptor will not change between the two setups (10 or 100 points per circles).

\subsection{Numerical comparison}

In this section we present a numerical comparison between the two methods. This comparison aims at illustrating the three following points:
\begin{enumerate}
    \item For the same energy, the gradient descent optimization method reaches low energy levels in less time (i.e. less evaluations of energy value) than the optimization method used in \citet{Tscheschel2006}. To highlight this point, we shall consider the Cox Voronoi example, and use the WPH descriptors (c.f. \eqref{eq:wphK}) to define the energy, and compare the RS and GD optimization methods.
    \item The amount of information captured by the $k$th nearest neighbours d.f.'s depends on the intensity of the process, regardless of the scales of the structures. Considering the Turbulent Poisson example with different intensities, we shall see that, for a fixed descriptor (i.e. fixed $k_{max}, \: r_{max}$, number of $r$), the quality of the syntheses decreases with the intensity of the process.
    \item We perform an overall comparison of the two methods, on the Cox Voronoi and Turbulence Poisson examples. Besides the visual and TDA comparison, we further provide statistical performance metrics to illustrate the better performance of our method.
\end{enumerate}

\subsubsection{Comparison between RS and GD (Cox Voronoi example)}


For this experiment, we study the Cox Voronoi example, and define the energy from the wavelet phase harmonics covariances, presented in Section \ref{s.Wavelets}. Let K be our descriptor (defined in \eqref{eq:wphK}), $\bp$ our observation sample, and $E_{\bp}(\cdot) = \frac{1}{2}| K (\cdot) - K(\bar{\phi}) |^2$ the corresponding energy. We define the relative energy by 

\begin{equation}
    e(\cdot) = 2\frac{ E_{\bar\Phi}(\cdot) } { | K(\bar{\phi})|^2 }  = 
    \frac{ | K (\cdot) - K(\bar{\phi}) |^2 } { | K(\bar{\phi})|^2 }.
\end{equation}

We ran the optimization of the energy with the random search method from \citet{Tscheschel2006}, and observe the relative energy of the syntheses (for 10 syntheses), after $n=19870$ and $n=29805$ iterations, i.e. respectively 10 and 15 iterations per point.
After $n=19870$ iterations, the algorithm reaches a relative energy of $e=9,00.10^{-4}$ (with a std of $9,23.10^{-5}$), and after $n=29805$ iterations, we found $e=4,76.10^{-4}$ (std$=2,47.10^{-5}$), indicating that the optimization has reached a low energy level. It took an average of 1h04min and 1h36min respectively. We observed the respective relative energies, and ran our gradient descent optimization algorithm (without the multi-scale procedure) until the relative energy reaches the levels from the random search method.
The results and comparisons with our method are summarized in Table \ref{tab:speed}.
The computations have been run on a single GPU Nvidia Tesla P100.

\begin{table}[!h]
    \centering
    \begin{tabular}{|c|c|c|}
    \hline
                    & Random search & Gradient descent \\
    \hline
        $e=9,00.10^{-4}$ &  19870 (1h04m) & 52 (0m35s)\\
    \hline
        $e=4,76.10^{-4}$ &  29805 (1h36m) & 69 (0m45s)\\
    \hline
    \end{tabular}
    \caption{Speed comparison between random search and gradient descent, in number of iterations (computation time in parenthesis) for the synthesis of Poisson Voronoi patterns. The time per iteration in the gradient descent method is larger, due to the possible several energy (and gradient) evaluations for the line search. However, the total amount of time is much lower.
}
    \label{tab:speed}
\end{table}


\subsubsection{Dependence of WPH and NND on the intensity of points (Turbulence Poisson example)}

To illustrate our second point, for NND descriptor we fix the parameters of the $k\,$th nearest neighbours d.f.'s to $k_{max}=16$, $r_{max}=.125$ (on a window of size 1), and discretize $D_k(r)$,  $r\in(0,r_{\max}]$, regularly by 250 values of  radii~$r$. With this fixed descriptor, we perform a synthesis using RS optimization for three different observations: the Turbulence Poisson observation randomly thinned to have 500 points, the same observation thinned to 2000 points, and the raw observation, which contains 3784 points. We set the number of iterations to 400 iterations per point, 
(which is the same for both methods).
Figure \ref{intensity} shows examples of syntheses for the 3 different patterns, as well as syntheses from our method using WPH descriptor with GD multiscale optimization. We observe that the method using the RS+NND fails to reproduce the geometric structures in the example with the largest number of points. 

\paragraph{Statistical evaluation metrics}

we present the estimations of two statistics.
The first one is the spherical contact distribution function (SCDF), defined for a point process $\Xi$ as  $H_s(r) := 1 - \mathbb P (\Xi \cap B(0,r) = \emptyset)$, where $B(0,r)$ denotes the ball of radius $r$, centered at 0.
The second one is the Euler-Poincar\'e characteristic, defined from the persistence diagram of a point pattern (cf. \ref{ss.TDA}) as the number of connected components minus the number of holes, in function of the radius $r$.
For these two statistics, each radius $r$, and each distribution, we estimate their value by averaging over 10 realizations. A confidence interval is computed using a bootstrap method, see e.g. \citet{efron1994introduction}, with 9999 resamples, a confidence level of .95, and the 'BCa' method. We also compare this estimation with the normalized standard deviation of our samples, under Gaussianity assumptions (see for instance \citet[Chapter 5]{efron1994introduction}).

The curves of Figure \ref{fig:thin-stats} confirm our visual evaluation: as the number of points in the pattern grows large, the error (deviation from the curve of the true distribution) becomes larger for the RS+NND model. In the example with the largest number of points, the SCDF curve of this model is significantly above the curve of the true distribution, because the observation contains clusters formed by a large number of points, which is not captured by the NND descriptor with $k_{max}=16$. Therefore, large empty regions are not reproduced, and the probability of having a point inside some ball of given radius is too high. Similarly, the curve of the Euler characteristic ($\chi$) of the RS+NND model deviates from the one of the true distribution when the number of points is large. 

To quantify more precisely these errors, we report in Table \ref{tab:hest-nothin} values of the two statistics for several relevant radii $r$.
{\setlength{\tabcolsep}{1.5pt}
\begin{figure}[!h]
\centering
\vspace{-2ex}
\begin{tabular}{rcc}
 & 500 points &  3784 points \\
\rotatebox[origin=l]{90}{\hspace{3em}Original}&
\includegraphics[width=0.45\linewidth]{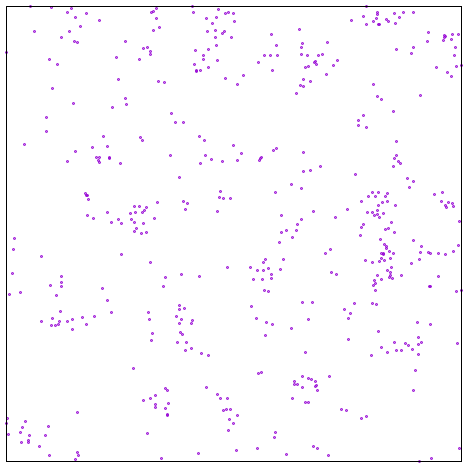}&
\includegraphics[width=0.45\linewidth]{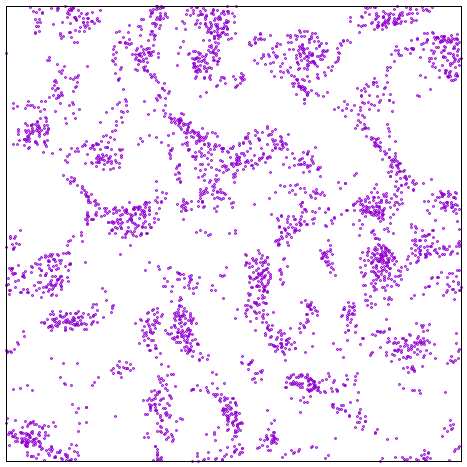}\\
\rotatebox[origin=l]{90}{\hspace{3em}RS+NND}&
\includegraphics[width=0.45\linewidth]{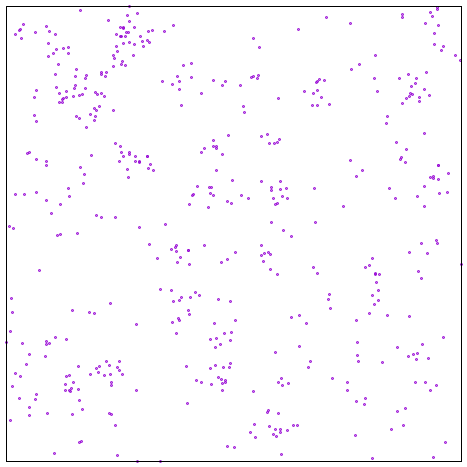}&
\includegraphics[width=0.45\linewidth]{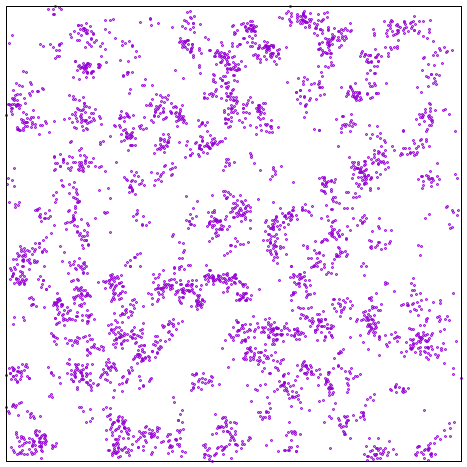}\\
\rotatebox[origin=l]{90}{\hspace{3em}WPH+GD}&
\includegraphics[width=0.45\linewidth]{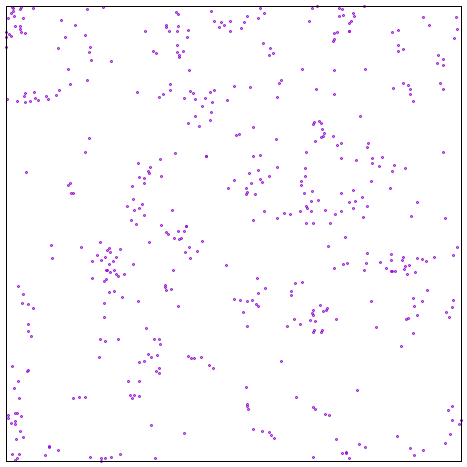}&
\includegraphics[width=0.45\linewidth]{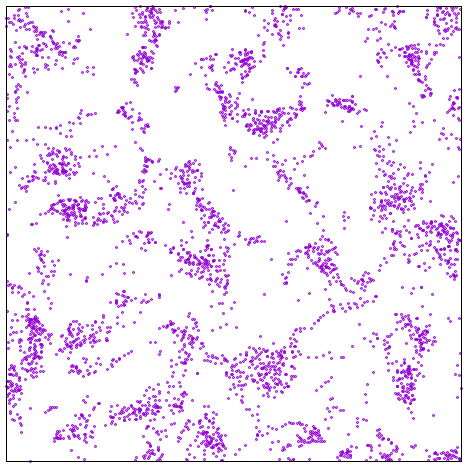}\\
\end{tabular}
\caption{Syntheses from the RS+NND model and our model, from observations containing 500 points (left) 3784 points (right).
} 
\label{intensity}
\end{figure}}

{\setlength{\tabcolsep}{1.4pt}
\begin{figure}[!h]
    \centering
    \begin{tabular}{rcc}
    & Thinning & No thinning\\
       \rotatebox[origin=l]{90}{\hspace{2em}SCDF}&
 \includegraphics[scale=.28]{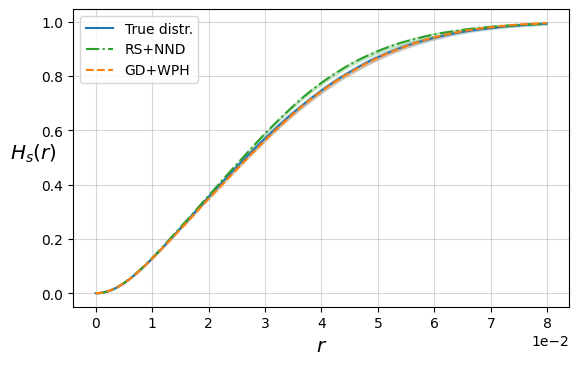}&
 \includegraphics[scale=.28]{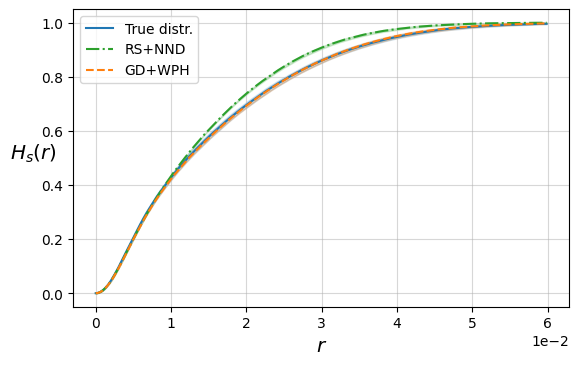}\\
 \rotatebox[origin=l]{90}{\hspace{2em}Euler}&
 \includegraphics[scale=.28]{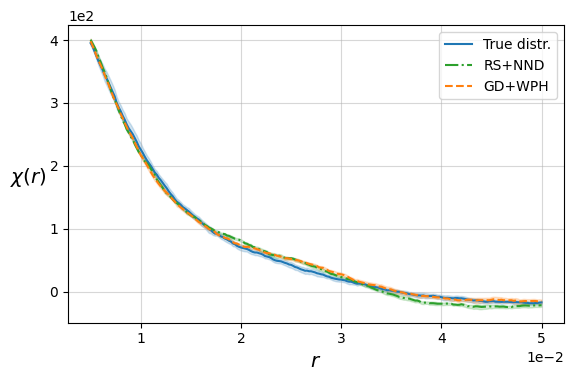}&
 \includegraphics[scale=.28]{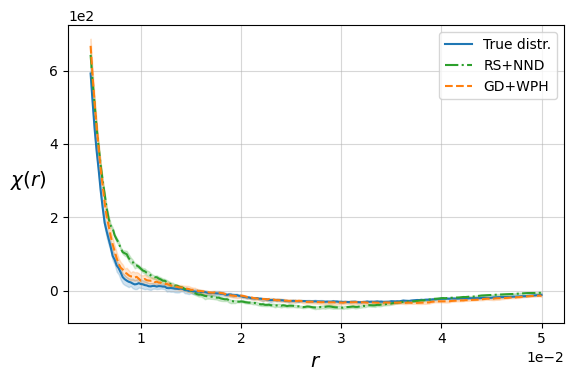}
    \end{tabular}
    \caption{Spherical contact distribution function (SCDF) and Euler-Poincar\'e characteristic, for the thinned and not thinned distributions. We compare the statistics of the true distribution, as well as the RS+NND and WPH+GD models.}
    \label{fig:thin-stats}
\end{figure}}

\begin{table*}[!h]
    \centering
    \begin{tabular}{|c|c|c|c|c|c|}
    & 1.5e-2 & 2.4e-2 & 3.3e-2 & 4.1e-2 & 5.0e-2 \\
True  & 5.8e-1(1.1e-2, 1.2e-2) & 7.7e-1(1.0e-2, 1.0e-2) & 8.9e-1(8.3e-3, 8.5e-3) & 9.5e-1(6.3e-3, 6.5e-3) & 9.8e-1(4.3e-3, 4.3e-3) \\
RS+NND & 6.0e-1(2.1e-3, 2.1e-3) & 8.1e-1(3.9e-3, 4.0e-3) & 9.3e-1(4.4e-3, 4.4e-3) & 9.8e-1(3.0e-3, 3.1e-3) & 1.0e+0(1.5e-3, 1.5e-3) \\
GD+WPH & 5.7e-1(4.2e-3, 4.2e-3) & \textbf{7.6e-1} (6.2e-3, 6.4e-3) & \textbf{8.9e-1}(6.7e-3, 6.9e-3) & \textbf{9.6e-1}(5.6e-3, 5.8e-3) & 9.9e-1(3.8e-3, 3.9e-3)

    \end{tabular}
    \caption{Comparison of the SCDF for Turbulence distributions without thinning, on a range of relevant values of $r$. In parenthesis, the half $95\%$ confidence interval, estimated with bootstrap and standard deviation respectively. Bold numbers indicate values significantly closer to the true distribution.}
    \label{tab:hest-nothin}
\end{table*}

\begin{table*}[!h]
    \centering
    \begin{tabular}{|c|c|c|c|c|c|}
    & 8.6e-3 & 1.6e-2 & 2.4e-2 & 3.2e-2 & 4.0e-2\\
True  & 2.7e+1(1.3e+1, 1.3e+1) & -4.1(5.8, 5.7) & -2.7e+1(2.3, 2.4) & -3.1e+1(2.8, 2.9) & -2.3e+1(1.7, 1.7) \\
RS+NND & 9.9e+1(8.2, 8.5) & -1.5e+1(6.1, 5.1) & -4.0e+1(4.0, 4.0) & -4.2e+1(3.1, 3.1) & -2.1e+1(1.4, 1.5) \\
GD+WPH & \textbf{4.6e+1}(1.4e+1, 1.4e+1) & -3.0(4.7, 4.5) & \textbf{-2.7e+1}(4.2, 4.2) & -3.2e+1(3.9, 3.9) & -2.9e+1(2.5, 2.5)
    \end{tabular}
    \caption{Comparison of the Euler characteristic for Turbulence distributions without thinning, on a range of relevant values of $r$. In parenthesis, the half $95\%$ confidence interval, estimated with bootstrap and standard deviation respectively. Bold numbers indicate values significantly closer to the true distribution.}
    \label{tab:euler-nothin}
\end{table*}

\subsubsection{Direct comparison}


On the two examples treated above, we shall illustrate the overall performance of both methods.
We ran 10 simulations of syntheses from the method of \citet{Tscheschel2006}, with $k_{max}=64$  for the Cox Voronoi example, and $k_{max}=128$ for the Turbulence Poisson example, with 400 iterations per point. For both models we perform also the syntheses from our multiscale gradient descent method (with 400 iterations). 

The two families of  examples of syntheses  share the same corresponding observation coming from the original distributions. We also simulate 9 other patterns for the original distributions, to compare the averaged statistics and the diversity among the original distribution and the models.

\begin{figure}[!h]
    \centering
    \begin{tabular}{cc}
         Cox Voronoi & Turbulence Poisson \\
         \includegraphics[width=0.46\linewidth]{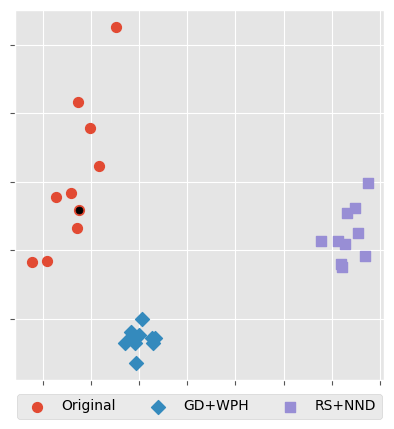} & \includegraphics[width=0.46\linewidth]{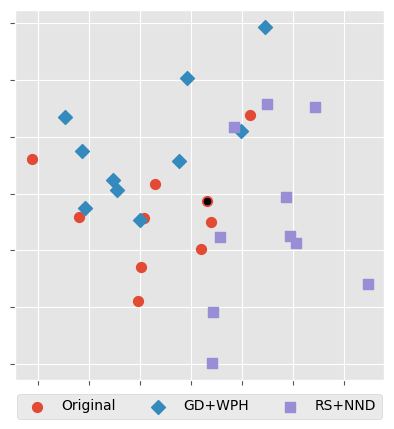}\\
         $d_{orig/RS+NND} = 1.52$ & $d_{orig/RS+NND} = 0.62$ \\
         $d_{orig/GD+WPH} = 0.75$ & $d_{orig/GD+WPH} = 0.61$\\
    \end{tabular}
    \caption{Visualization of the TDA of the three distributions (Original, GD+WPH, RS+NND), for the Cox Voronoi example (left), and the Turbulence Poisson example (right). The black point represents the observation pattern used for the syntheses. The  averaged  (true) distances between the original and synthesized patterns (via Wasserstein distance of persistence diagrams)  are given as well.}
    \label{DC_TDA}
\end{figure}

We compare the different distributions with visual evaluation (Figure \ref{DC}), TDA (Figure \ref{DC_TDA}), and by estimation of the SCDF and Euler characteristic (Figure \ref{DC_H}, Tables \ref{tab:hest-turb} and \ref{tab:euler-vor}). 
For the evaluation with TDA, in addition to the visualization of a 2-dimensional representation of the distance matrix between (the PD of) all original and synthesized point patterns (cf. Section \ref{ss.TDA}), we also compute the average Wasserstein distance between all pairs of point patterns belonging to different distributions. In more details, for a given distribution (Cox Voronoi or Turbulence Poisson), let $M:=M_{orig/GD+WPH}$ be the 10$\times$10 distance matrix 
between the 10 realizations of the original distribution and the 10 realizations of our model. We compute $d_{orig/GD+WPH} = \frac{1}{100}\sum_{i,j}M_{i,j}$.
Similarly, we compute   $d_{orig/RS+NND}$ for the RS+NND model. We obtained, for the Cox Voronoi example, $d_{orig/GD+WPH} = 0.75$, and $d_{orig/RS+NND} = 1.52$, showing a significant advantage to our method. Our experiments on the Turbulence Poisson example gave $d_{orig/GD+WPH} = 0.61$, and $d_{orig/RS+NND}\allowbreak = 0.62$. These results are coherent with the visual evaluation, that indicates a better performance of our model, particularly for the Cox Voronoi example.

Figure \ref{DC_H} also confirms our visual evaluation. The curve of the $\chi$ function clearly shows a larger error for the RS+NND model. Indeed, we can observe from the alignment of points in the observation that the number of connected components quickly decreases in the patterns of the true distribution. As this alignment is not as well reproduced in the RS+NND model as in ours, we observe that the curve of the $\chi$ function decreases more slowly for the RS+NND model. Additionally, the SCDF curves for the Turbulence example show a significant error for the RS+NND model, for which the curve is above the true distribution curve, indicating the presence of fewer large empty areas around clusters.
For the Voronoi example however, our model also shows a significant error on the SCDF curve, similar to the RS+NND model, possibly due to the presence of points inside the formed cells (only one is needed to impede the presence of an empty region or a hole). This can also explain the error observe on the TDA plot of the Voronoi distributions (Figure \ref{DC_TDA}, left).

{\setlength{\tabcolsep}{1.5pt}
\begin{figure}[!h]
\centering
\vspace{-2ex}
\begin{tabular}{ccc}
\rotatebox[origin=l]{90}{\hspace{3em}Original}&
\includegraphics[width=0.45\linewidth]{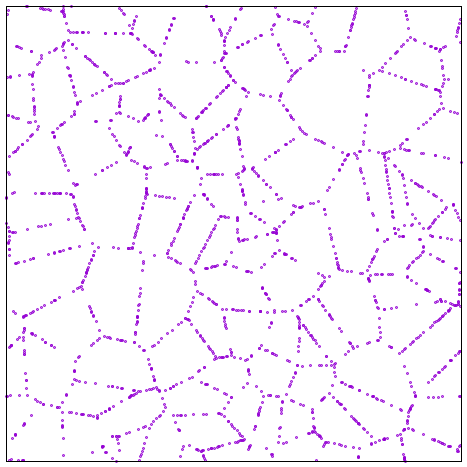}&
\includegraphics[width=0.45\linewidth]{Figures/comp_stoyan/obs-nothin.png}\\
\rotatebox[origin=l]{90}{\hspace{3em}RS+NND}&
\includegraphics[width=0.45\linewidth]{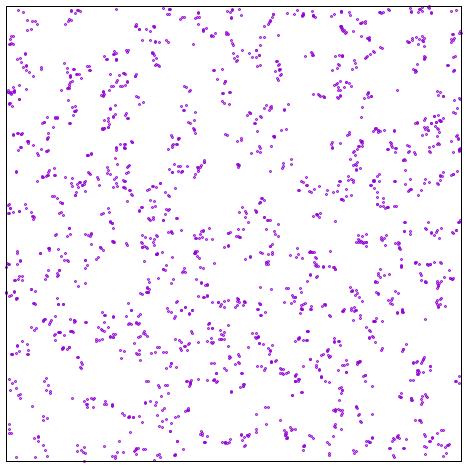}&
\includegraphics[width=0.45\linewidth]{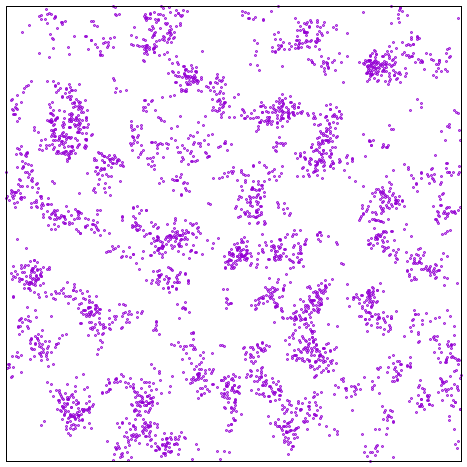}\\
\rotatebox[origin=l]{90}{\hspace{3em}GD+WPH}&
\includegraphics[width=0.45\linewidth]{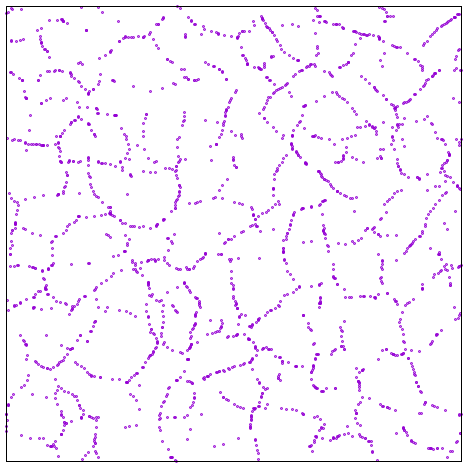}&
\includegraphics[width=0.45\linewidth]{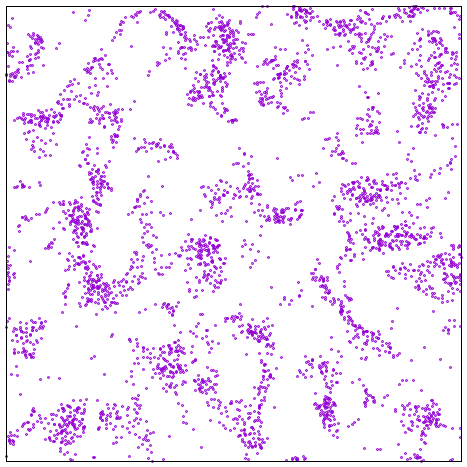}\\
\end{tabular}
\caption{
Visualization of syntheses for the Cox Voronoi and Turbulence Poisson examples. Top: observations, middle: RS+NND, bottom: GD+WPH.
}
\label{DC}
\end{figure}}

{\setlength{\tabcolsep}{1.5pt}
\begin{figure*}[!h]
    \centering
    \begin{tabular}{rcc}
    & Cox Voronoi & Turbulence Poisson\\
    \rotatebox[origin=l]{90}{\hspace{5em}SCDF}&
    \includegraphics[scale=.4]{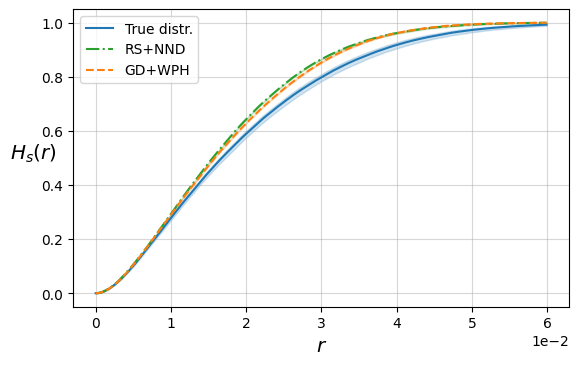}&
    \includegraphics[scale=.4]{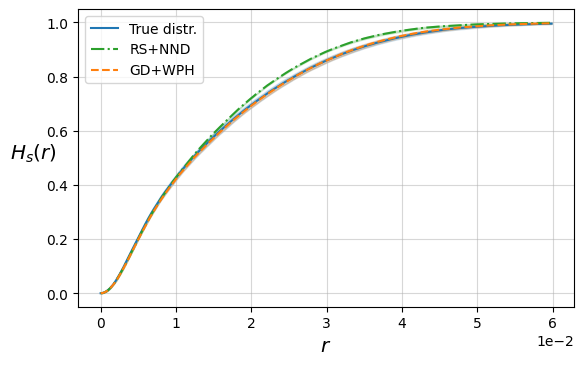}\\
    \rotatebox[origin=l]{90}{\hspace{5em}Euler}&
    \includegraphics[scale=.4]{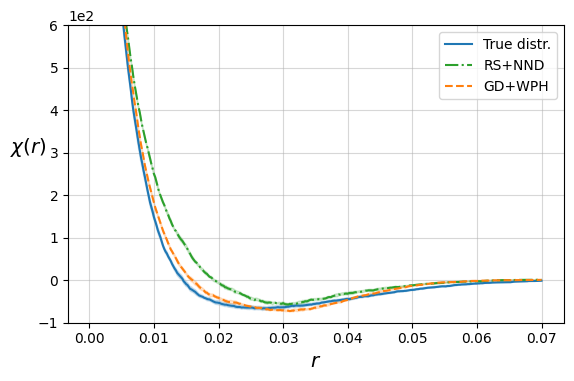}&
    \includegraphics[scale=.4]{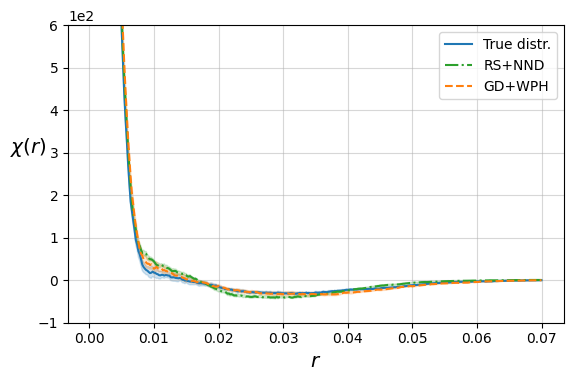}
    \end{tabular}
    \caption{Spherical contact distribution function (SCDF) and Euler-Poincar\'e characteristic, for the Voronoi and Turbulence distributions. We compare the statistics of the true distributions, the RS+NND, and the WPH+GD models.}
    \label{DC_H}
\end{figure*}}


\begin{table*}[!h]
    \centering
    \begin{tabular}{|c|c|c|c|c|c|}
    & 1.5e-2 & 2.3e-2 & 3.0e-2 & 3.7e-2 & 4.5e-2 \\
True  & 5.8e-1(1.1e-2, 1.2e-2) & 7.5e-1(1.0e-2, 1.1e-2) & 8.6e-1(8.8e-3, 8.9e-3) & 9.3e-1(7.1e-3, 7.3e-3) & 9.7e-1(5.5e-3, 5.6e-3) \\
RS+NND & 5.9e-1(1.4e-3, 1.4e-3) & 7.8e-1(1.8e-3, 1.8e-3) & 8.9e-1(2.5e-3, 2.6e-3) & 9.6e-1(3.3e-3, 3.3e-3) & 9.9e-1(2.9e-3, 3.0e-3) \\
GD+WPH & 5.7e-1(4.2e-3, 4.2e-3) & 7.4e-1(6.1e-3, 6.3e-3) & \textbf{8.6e-1}(7.0e-3, 7.1e-3) & \textbf{9.3e-1}(6.2e-3, 6.3e-3) & 9.7e-1(4.8e-3, 4.9e-3)

    \end{tabular}
    \caption{Comparison of the SCDF for Turbulence distributions, on a range of relevant values of $r$. In parenthesis, the half $95\%$ confidence interval, estimated with bootstrap and standard deviation respectively. Bold numbers indicate values significantly closer to the true distribution.}
    \label{tab:hest-turb}
\end{table*}

\begin{table*}[!h]
    \centering
    \begin{tabular}{|c|c|c|c|c|c|}
    & 1.0e-2 & 2.0e-2 & 3.0e-2 & 4.0e-2 & 5.0e-2 \\
True  & 1.5e+2(7.2, 7.4) & -5.3e+1(5.0, 5.2) & -6.4e+1(4.2, 4.3) & -4.4e+1(3.2, 3.2) & -2.3e+1(1.7, 1.7) \\
RS+NND & 2.5e+2(3.9, 3.7) & -6.5(2.1, 2.1) & -5.4e+1(3.3, 3.4) & -3.2e+1(3.0, 3.1) & -1.2e+1(1.2, 1.2) \\
GD+WPH & \textbf{1.8e+2}(3.1, 3.1) & \textbf{-4.1e+1}(3.2, 3.3) & -7.1e+1(2.1, 2.2) & -4.6e+1(1.9, 1.9) & -1.3e+1(8.5e-1, 8.6e-1)
    \end{tabular}
    \caption{Comparison of the Euler characteristic for Voronoi distributions, on a range of relevant values of $r$. In parenthesis, the half $95\%$ confidence interval, estimated with bootstrap and standard deviation respectively. Bold numbers indicate values significantly closer to the true distribution.}
    \label{tab:euler-vor}
\end{table*}


\section{Conclusion}

In this paper, we present a particle gradient descent model 
to simulate stationary and ergodic point processes, based on a single observation in a square window. 
This model is able to synthesize processes formed by a large number of points, exhibiting interactions at multiple scales.
Our method is built upon recent works on gradient descent methods 
to approximate the micro-canonical model. 
To characterize complex geometric point patterns, 
we use the wavelet phase harmonic descriptors that allow to 
explicitly control the scales of the structures to model.
Numerical results on Cox and Turbulent distributions 
validate the ability of the model to capture various
geometric structures in the observation.
Compared to the classical approaches developed in \citet{torquato, Tscheschel2006},
our approach brought a new perspective 
to the modeling of point processes,
through the lens of wavelet analysis and image modeling. 

\section*{Conflict of interest}

The authors declare that they have no conflict of interest.

\bibliographystyle{spbasic}
\bibliography{generative2}

\appendix

\section{Proof of Theorem~\ref{th.Invariance}}
\label{s.Invriance-proof}

In order to prove Theorem~\ref{th.Invariance}, we need to formally define Eq.~\eqref{e.particle-gradient}. Recall that  in this section and in what follows,  $W_s$ is interpreted as endowed
with the addition and scalar multiplication 
modulo $W_s$. 

For $\mu \in  \M^s$ and any $x \in \text{Supp}(\mu)$, we define the following functions:
\[
\begin{array}{ccccc}
{h_{x}^{\mu}}& : & \mathbb R^2 & \longrightarrow & \mathbb M^s \\
 & & y & \longmapsto & \displaystyle{\mu - \delta_x + \delta_{x+y},} \\
\end{array}
\]
\[
\begin{array}{ccccc}
K_{x}^{\mu}& : & \mathbb R^2 & \longrightarrow & \C^d \equiv \R^{2d} \\
 & & y & \longmapsto & \displaystyle{K\circ h_{x}^{\mu}(y)}, \\
\end{array}
\]
\[
\begin{array}{ccccc}
E_{x}^{\mu}& : & \mathbb R^2 & \longrightarrow & \mathbb R^+ \\
 & & y & \longmapsto & \displaystyle{E_{\bp}\circ h_{x}^{\mu}(y)}.
\end{array}
\]

The function $K_{x}^{\mu}$ can be complex valued. However, as our energy function is the square Euclidean norm, it is equivalent to consider that $K_{x}^{\mu}$ has values in $\R^{2d}$. Moreover, we assume in what follows that the  function $K$ is such that for all $\mu \in \M^s$ and all  $x \in \text{Supp}(\mu)$, $K_x^\mu$ is differentiable.
We can then define from chain rule, for any $\mu \in \M^s$ and any $x \in W_s$ 
\begin{align}
\nabla_xK(\mu) := 
\left\{
    \begin{array}{ll}
        Jac[K_x^\mu](0) & \mbox{if } x \in \text{Supp}(\mu)\\
        0 & \mbox{otherwise}  
    \end{array}
\right.,\\
\noalign{and}\nonumber\\
\nabla_xE_{\bp}(\mu) := \left\{
    \begin{array}{ll}
        Jac[E_x^\mu](0)  & \mbox{if } x \in \text{Supp}(\mu)\\
        0 & \mbox{otherwise} 
    \end{array}
\right.,
\label{e.NablaE}
\end{align}
where $Jac[f]$ denotes the Jacobian matrix of the function $f$.
When $x \in \text{Supp}(\mu)$, the chain-rule gives $Jac[E_x^\mu](0) = (\nabla_xK(\mu))^t(K(\mu)-K(\bp))$. We can now give the proof of Theorem~\ref{th.Invariance}.

\begin{proof}
    We are going to show that if $\Phi_n$ follows a distribution invariant to $T$, 
    then $\Phi_{n+1} = G_{\bp} (\Phi_n)$ also follows a distribution that is invariant to $T$. 
    The gradient descent procedure thus produces a sequences of measures $\Phi_n$ that are all invariant to $T$ 
    because the initial random measure $\Phi_0$ is invariant invariant to $T$. 
    
    Denote $G_{\bp} (\mu)$ by the measure configuration transported from $\mu$, by performing one gradient-descent step on the energy $E_{\bp}$. More precisely, for $\mu = \sum_i \delta_{x_i}$, we define for a fixed $\gamma>0$, the gradient-descent step by 
    \[
        G_{\bp} (\mu) := \sum_i \delta_{x_i - \gamma \nabla_{x_i} E_{\bp} (\mu) } .  
    \]
    For any transform $T x = Ax + b$ on $\W$, where $A$ is an orthogonal matrix $A$ with entries in $\{-1,0,1\}$, 
    and $b\in \W$. As $A$ is a linear transformation on the torus $\W$, $\forall x , y \in \W$, $A(x+y) = Ax + Ay$.     
    We shall first prove that 
    \begin{equation}
        G_{\bp} (T_{\#} \mu) = T_{\#} G_{\bp} ( \mu). 
        \label{onestep}
    \end{equation}
    Let  $y_i = T x_i$, then by definition,
    \[
         G_{\bp} ( T_{\#}\mu) = \sum_i \delta_{ y_i - \gamma \nabla_{y_i} E_{\bar \mu} (T_{\#} \mu ) } ,
    \]
    and
    \[
         T_{\#} G_{\bp} ( \mu) = \sum_i \delta_{T( x_i - \gamma \nabla_{x_i} E_{\bp } (\mu ) ) }. 
    \]
    We are going to show that for each $i$-th particle, $y_i - \gamma \nabla_{y_i} E_{\bp} (T_{\#} \mu ) = T( x_i - \gamma \nabla_{x_i} E_{\bp } (\mu ) )$. This implies that \eqref{onestep} is correct. 
    The key is to show that 
    \begin{equation}
           A \nabla_{x_i} E_{\bp} (\mu ) = \nabla_{y_i} E_{\bp} (T_{\#} \mu ) 
           \label{eq:key}
    \end{equation}
    which will imply that $\forall i$,
    \[
        T( x_i - \gamma \nabla_{x_i} E_{\bp } (\mu )  ) = A x_i - \gamma A \nabla_{x_i} , 
    \]
    \[
        E_{\bp} (\mu ) + b = y_i - \gamma A \nabla_{x_i} E_{\bp} (\mu ) = 
        y_i - \gamma \nabla_{y_i} E_{\bp} (T_{\#} \mu ) .
    \]
    To show \eqref{eq:key}, we recall 
    that by the definitions in Section \ref{ss.Gradient-descent},  
    \begin{align}
        &\nabla_{x_i} E_{\bp} (\mu ) = Jac(K_{x_i}^\mu)(0)^t ( K ( \mu) - K(\bp) ) , \label{eq:jac1}       \\ 
        &\nabla_{y_i} E_{\bp} (T_{\#} \mu ) = Jac(K_{y_i}^{T_{\#} \mu })(0)^t ( K (T_{\#} \mu) - K(\bp) ) ,
        \label{eq:jac2}        
    \end{align}
    with
    $Jac(K_{y_i}^{T_{\#} \mu })(0) = Jac(K_{y_i}^{T_{\#} \mu } \circ A \circ A^{-1})(0) = Jac(K\circ h_{y_i}^{T_{\#}\mu} \circ A)(0)A^{-1}.$  
    Furthermore, $\forall x \in \W$,
    \begin{align}
      K\circ h_{y_i}^{T_{\#}\mu} \circ A (x) &= K(T_{\#}\mu - \delta_{y_i} + \delta_{y_i + Ax}) = K(T_{\#}(\mu - \delta_{x_i} + \delta_{x_i + x})) \nonumber \\
      &= K(\mu - \delta_{x_i} + \delta_{x_i + x}) = K \circ h_{x_i}^\mu(x),
      \label{eq:jac3}  
    \end{align}
    where we used the fact that $T$ is affine, and the invariance of $K$ w.r.t. $T$.
    The equality in \eqref{eq:key}  follows directly from \eqref{eq:jac1},\eqref{eq:jac2},\eqref{eq:jac3} and the fact that $A^{-1} = A^t$. From \eqref{eq:key}, we conclude that \eqref{onestep} holds. 
    
    Based on \eqref{onestep}, it remains to show that for any Borel set $\Gamma $ on $\W$, $ P ( \Phi_n \in T_{\#}^{-1} \Gamma ) = P ( \Phi_n \in \Gamma )$. This can be shown by induction, since by assumption it holds at $n=0$: assume now that this statement holds at $n \geq 0$, then we have, 
    \begin{align}
    P ( \Phi_{n+1} \in T_{\#}^{-1} \Gamma )
    &= P (  G_{\bp} (\Phi_n) \in T_{\#}^{-1} \Gamma )\\ \notag
    &= P (  T_{\# } G_{\bp} (\Phi_n) \in  \Gamma ) \\ \notag
    &= P( G_{\bp} ( T_{\# } \Phi_n) \in  \Gamma   ) \\ \notag
    &= P( G_{\bp} (  \Phi_n) \in  \Gamma   )\\ 
    &= P ( \Phi_{n+1} \in \Gamma ) \label{els}.
    \end{align}
    The second last equality in \eqref{els} is due to the invariance of $\Phi_n$, i.e. 
    \begin{align*} 
    P( G_{\bp} ( T_{\# } \Phi_n) \in  \Gamma   ) &= P ( \Phi_n \in T_{\#}^{-1}  G_{\bp}^{-1} \Gamma )\\
    &= P ( \Phi_n \in G_{\bp}^{-1} \Gamma )\\
    &= P( G_{\bp} (  \Phi_n) \in  \Gamma   ).
    \end{align*}

\end{proof}

\section{Fourier spectrum and power spectrum}

\label{s.Fourier}

We define the  {\em discrete Fourier transform} (DFT) $F_m(\mu)$ of a counting measure $\mu =\sum_u\delta_{x_u}\in\M^s$ on 
the (square) window $[-s,s[^2$
at integer frequency $m \in \Z^2 $ by
\[
    F_m(\mu):=\int_{\W} e^{-i\pi mx/s}\,\mu(dx)=\sum_{u} e^{-i\pi m x_u/s}.
\]
Observe, $F_m(\mu)$ at frequency $m=(0,0)$ specifies the number of points of the measure $\mu $ on $\W$. 
The empirical {\em Fourier spectrum} (or {\em power spectrum }) is often defined by taking the square modulus of the Fourier coefficients $F_m(\mu)$; $U_m(\mu):=|F_m(\mu)|^2 $. Note that $|F_m(\mu)|^2$, and consequently $U_m(\mu)$ is invariant with respect to (circular) translations of $\mu$ on $W_s$.
By selecting the  frequencies in a limited range $m \in \Gamma_F \subset \Z^2$, one obtains 
a translation-invariant Fourier spectrum. As we shall focus on isotropic point processes, we further reduce the variance of our statistics by averaging Fourier coefficients along frequency orientations. More precisely, let us define $\tilde\Gamma_F := \{ \lfloor |m| \rfloor, \: m \in \Gamma_F \}$. For each $k\in \tilde \Gamma_F$, we define 
$\tilde U_k(\mu) := \frac{1}{\#k}\sum_{\substack{m \in \Gamma_F \\ \lfloor |m| \rfloor = k}} U_m(\mu)$, where $\#k$ denotes the cardinal of $\{ m \in \Gamma_F : \lfloor |m| \rfloor = k \}$.
The radial power spectrum $P(k)$ is the expectation of $\tilde U_k(\mu) $ for $k \in N = \{ 1,2,3,... \} $ when $\mu$ follows some distribution, divided by the intensity of the process (estimated over 10 realizations).

\section{Relaxing the assumptions on the data}\label{s.relax}

In this paper, in order to present our model in a simple setting, strong theoretical assumptions have been made on the data. However, in real world applications, the data will most likely not satisfy these assumptions. This sections presents ideas on how to adapt our model in such cases.

\paragraph{Non-periodic boundaries} Recall that our descriptor, defined in \eqref{eq:wphK}, applies periodic boundary correction to point patterns in a square window. If the structure of the observed pattern is not periodic, one can modify the descriptor by applying non-periodic integrals in~\eqref{eq:wphK} over some smaller window. In particular, we suggest  a  {\em scale-dependent reduction of  the integration window}, pertinent when the wavelet $\psi$ has a compact (or approximately compact) spatial support. Specifically, we consider a new descriptor $\tilde K$  by considering the integrals in~\eqref{eq:wphK} with $i=(\la,k,\la',k', \tau') \in \Gaw$ 
over smaller windows  $W_{s_i}\subset W_s$, such that boundary effects are negligible. Our current software can also handle such non-periodic boundary conditions.

\paragraph{More general observation windows} In this paper, we considered that the observed pattern lies in a square observation window. If this is not the case, one could use a similar idea to the non-periodic case: embed the observation window in a square window and considering integrals in~\eqref{eq:wphK} over the observation window.

\paragraph{Non stationary process} In \citet{ppsim}, the authors focus on building a model for non stationary point processes inspired by \citet{Tscheschel2006}. Similarly, one might adapt our method to model non stationary processes. This could be done by modifying two aspects of the method. First, the initial distribution $\Phi_0$ (cf. Section \ref{ss.Gradient-descent}) could be chosen as a non stationary Poisson point process, estimating the intensity with a kernel estimator, such as in \citet{ppsim}. In addition, as pointed out in \citet{ppsim}, the descriptor should be adapted not to be translation invariant. This could be done, for example, by applying local integrals over patches of the observation window in~\eqref{eq:wphK} (this requires some notion of "local stationarity" of the process). Another method that may be useful in this scenario is the regularization proposed in \citet{brochard2020particle}, where a regularization term is added to the energy. This term consists of the (Sliced Wasserstein) distance \citep{rabin2011wasserstein} between the initial configuration and the current configuration (the one being optimized). By adding this regularization term to the energy, the points of the configuration are forced not to move too far away from the initial configuration, which could help preserve the non stationarity of the initial distribution in the distribution of the model.

\paragraph{Processes in other dimensions}
While we focus in this paper on planar point processes, our approach can readily be extended to any dimensions. 
To model point processes in other dimensions such as 1d or 3d, 
one can consider similar type of wavelets proposed in the literature 
\citep{5872834,brumwell_steerable_2018}.

\section{List of important parameters of our model}\label{s.params}

In Table \ref{tab:param}, we discuss the main parameters of our model in three categories. The first two categories are the parameters that are relatively standard to consider in most existing methods such as \citet{Tscheschel2006}. The third category is more specific to our model, which involves the discretization step, and the final blurring step. 
 
\begin{table*}[h!]
    \begin{tabularx}{\textwidth}{|c|c|X|}
        \hline
        Category & Parameter &  Discussion \\
        \hline 
        \hline
        \textbf{Descriptor}   (c.f. Section \ref{gamma_choice} and \ref{sss:waveletdisc}) & Number of wavelet scales $J$ & 
        The scales of the wavelet transform are defined by $0 \leq j < J$.
        The minimal scale $j=0$ is 
        chosen though the image resolution $N$ of the discretization,
        and is related to 
        the precision in high frequencies. 
        This choice should depend on the observed pattern. The maximal scale $J$ should be as large as possible, to capture enough structural information, but not too large, so that the wavelet phase harmonic
        covariances remain empirically well estimated.
        \\ 
        & Number of wavelet orientations $L$ & 
        This determines the angular precision of the descriptor. 
        A larger $L$ captures finer orientations of edge-like structures.  \\
        & Range of phase harmonics $(k,k')$   & 
        This determines the range of interactions between the wavelet phase harmonics coefficients. 
        The choice of $(k,k')=(1,1)$ corresponds to the second order statistics. \\
        \hline
        \hline
        \textbf{Optimisation }  
        (cf Section \ref{ss.Gradient-descent} and \ref{ss.it}) & Number of iterations &
        In our experiments, we set a fixed number of iterations.
        We found that increasing the number of iterations further only decrease the energy of the configurations by a small factor. 
        Other standard stopping criteria based on the norm of gradient can also be considered.   \\
        \hline
        \hline
        \textbf{Extra steps}
        (cf. Section \ref{s.Scheme}) & Image resolution $N$ & The larger the resolution, the smaller the structures of point processes which we can model. 
        However, there is an extra computational cost when $N$ increases. It also results in a larger number of moments to estimate. \\
        & With or without multi-scale optimization  &  Multi-scale optimization has been found useful in the case where the maximal scale $J$ is large, to avoid poor local minima and reconstruct the observation. We have also used it in our synthesis experiments, to reduce the energy of the syntheses.
        \\ 
        & Final blurring & The final blurring is useful in the cases where the number of points per pixel is often larger than 1, to remove artifacts due to the discretization. \\ 
        \hline
    \end{tabularx}    
    \caption{Discussion of the important parameters of our model.}
    \label{tab:param}
\end{table*}
\end{document}